\documentclass{article}

 \usepackage[preprint]{neurips_2025}

\usepackage[utf8]{inputenc} 
\usepackage[T1]{fontenc}    
\usepackage{hyperref}       
\usepackage{url}            
\usepackage{booktabs}       
\usepackage{amsfonts}       
\usepackage{nicefrac}       
\usepackage{microtype}      
\usepackage{xcolor}         

\usepackage{graphicx} 

\usepackage{algorithm}
\usepackage{algorithmic}
\usepackage{amssymb}

\usepackage{listings}

\usepackage{longtable}

\title{Scaling LLM Planning: NL2FLow \\for Parametric Problem Generation and Rigorous Evaluation}

\author{%
  Jungkoo Kang \\ 
  IBM Research\\
  MIT-IBM Watson AI Lab\\
  314 Main St \\
  Cambridge, MA 02142 \\
  \texttt{jungkoo.kang@gmail.com} \\
}

\begin{document}

\maketitle

\begin{abstract}
Robust workflow composition is critical for effective agent performance, yet progress in Large Language Model (LLM) planning and reasoning is hindered by a scarcity of scalable evaluation data. This work introduces NL2Flow, a fully automated pipeline for generating and evaluating workflow planning problems. NL2Flow generates problems parametrically in a structured intermediate representation, translating them into both natural language and formal PDDL. I evaluate several open-source, instruct-tuned LLMs on a dataset of 2296 low-difficulty problems generated by NL2Flow. Results demonstrate that the best-performing model achieved 86\% success in generating valid plans and 69\% in generating optimal plans (for solvable problems). Regression analysis shows that the influence of problem characteristics on plan generation is contingent on both model and prompt design. Importantly, translating natural language problems into a structured JSON representation prior to symbolic planning significantly improved success rates, suggesting a benefit from neuro-symbolic integration. These findings underscore the importance of understanding error sources within LLM reasoning as systems scale to more complex tasks. As LLM reasoning scales to increasingly complex problems, understanding the shifting bottlenecks and sources of error within these systems will be crucial.
\end{abstract}

\section{Introduction}

Agentic systems, increasingly deployed to automate complex business processes, rely on the ability to dynamically compose tools and agents into effective workflows. The core challenge in building these systems is planning: determining the optimal sequence of actions to achieve a given goal. While traditional AI planning techniques offer formal guarantees, they struggle with the scalability and adaptability demanded by real-world applications involving a large number of tools and rapidly changing environments.

\begin{figure}
   \centering
  \includegraphics[scale=0.4]{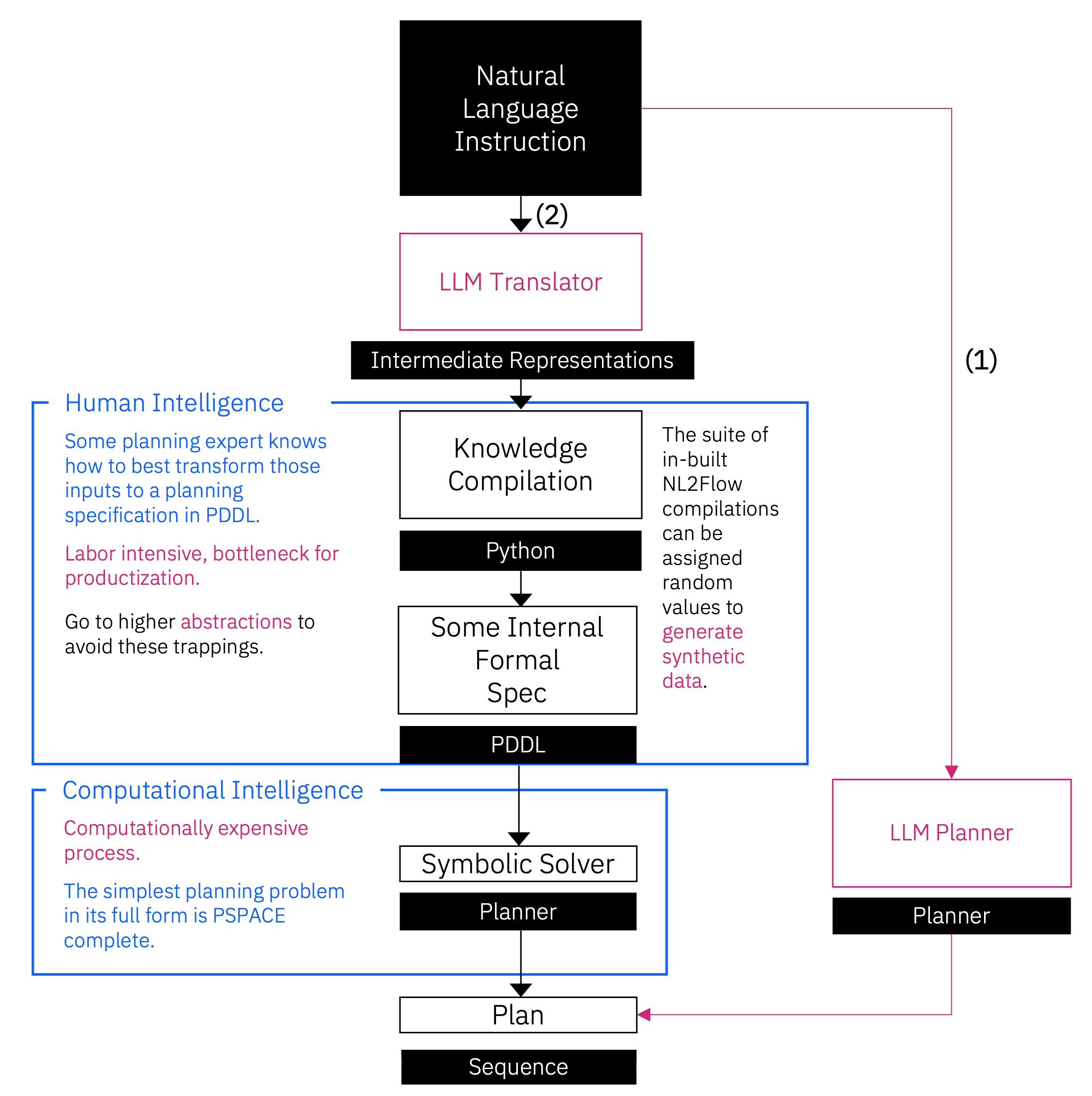}
  \caption{Steps to obtain a plan from a planning problem in natural language. Two approaches are shown: (1) Direct planning, where a Large Language Model (LLM) Planner directly generates a plan from the input. (2) Intermediate representation planning, where an LLM Translator first converts the natural language instruction into an intermediate representation, which is then used by symbolic planning algorithms to compute a plan.}
  \label{fig:overview}
\end{figure}

Traditional AI planning systems rely on formal representations, such as those defined by Planning Domain Definition Language (PDDL). While these systems can offer guarantees like optimality, building them requires painstaking knowledge engineering to translate natural language goals into formal problem descriptions \citep{georgievski2015htn}. This translation process is a known bottleneck, and the complexity of maintaining and scaling these systems increases substantially with the number of tools and task complexities. For example, business automation systems can involve thousands of APIs. Consequently, symbolic planners struggle to find optimal workflows within reasonable timeframes, often requiring developers to manually reduce the scope of the problem by limiting the number of APIs considered.

Large Language Models (LLMs) present a promising alternative, demonstrating impressive capabilities in natural language understanding, generation, and reasoning. However, using LLMs for reliable planning remains challenging. Existing approaches often resort to free-form plan generation, which lacks logical consistency, or rely on reinforcement learning confined to limited environments \citep{planningopen}. A key advancement would be the ability to reliably translate natural language planning problems into formal representations or to directly generate logically sound and executable plans (Figure \ref{fig:overview}). Successful LLM-based planning would significantly reduce the engineering effort associated with building robust natural language planning systems.

Despite their potential, evaluating and improving LLM planning capabilities is hampered by the scarcity of scalable data generation and evaluation methodologies. This study addresses this need by introducing NL2Flow (\url{https://github.com/IBM/nl2flow}), a scalable data generation and evaluation method for workflows. NL2Flow uses random and parametric approaches to create a dataset of workflow problems, eliminating the need for human or LLM annotation and enabling the creation of large-scale datasets. I demonstrate that NL2Flow can generate workflow problems in a domain where open-sourced, instructed-tuned LLMs without task-specific optimization or architectural modifications exhibit promising planning performance for low-difficulty problems. Establishing a domain in which large language models (LLMs) exhibit promising reasoning and planning capabilities without task-specific optimization and modifications can enable researchers to investigate the causes of their domain-dependent reasoning/planning performance.

Importantly, the use of PDDL allows for symbolic evaluation of generated plans, providing mathematically grounded metrics, which include optimality, for reliable and scalable assessment. By systematically varying goals and agent/tool configurations, NL2Flow facilitates evaluation of LLMs both as planners and as translators from natural language to workflow representations, highlighting key bottlenecks in building robust, natural language-grounded AI planning systems (Figure \ref{fig:overview}). In this study, I focus on applying LLMs to low-difficulty workflow generation problems, which are readily solvable by symbolic planners, as a first step towards tackling more complex scenarios.

\section{Method}

This study used an automated planning integration system, NL2Flow, to generate a dataset of workflow planning problems and evaluate the performance of large language models in generating workflows and translating workflow problems in natural language. The NL2Flow's profiler module (profiler.generators.batch\_data\_generator) was used to programmatically generate a diverse set of planning scenarios, varying parameters. This allowed for controlled generation of data suitable for evaluating LLM performance across a range of workflow configurations. The code for running experiments is available in the NL2FLOW-Runner repository (\url{https://github.com/IBM/nl2flow-runner}). LLM inference was conducted on compute clusters equipped with NVIDIA H100, H200, and A100 GPUs (with larger models, for example, DeepSeek-V3, using up to 8 H200 GPUs with 400GB memory). LLM-generated workflows were evaluated using NL2Flow’s Debugger module (nl2flow.debug.debug.BasicDebugger), which assesses soundness, validity, and optimality. A description of the AI planning principles is provided in the Appendix.

\subsection{Domain Description}

\begin{center}
\begin{lstlisting}[basicstyle=\small, frame=tb, breaklines=true, numbers=none, label={lst:nl2flowplanexample}, caption={NL2Flow domain example for booking a business trip (\url{https://github.com/IBM/nl2flow})}]
[0] ask(conference name)
[1] ask(username)
[2] invoice = Registration Bot(conference name, username)
[3] map(invoice, conference registration)
[4] name, address, Employee ID, Passport = Travel Info Agent(username)
[5] visa = Visa Application(Passport, address)
[6] ask(start date)
[7] ask(end date)
[8] map(home, address)
[9] map(BOS, destination)
[10] booking = Taxi(date, address, destination)
[11] map(LAX, address)
[12] map(JW Marriott Los Angeles LA 900 W Olympic Blvd, destination)
[13] booking = Taxi(date, address, destination)
[14] map(end date, date)
[15] map(JW Marriott Los Angeles LA 900 W Olympic Blvd, address)
[16] map(LAX, destination)
[17] booking = Taxi(date, address, destination)
[18] map(BOS, address)
[19] map(destination, home)
[20] booking = Taxi(date, address, destination)
[21] flight_ticket, hotel_booking = Concur(start date, end date, home, destination)
[22] map(flight_ticket, ticket to conference)
[23] assert $hotel_booking.price + $flight_ticket.price < 1500
[24] approval = Trip Approval(ticket to conference, conference registration)
\end{lstlisting}
\end{center}

Automated workflow completion was a core requirement for the intelligent systems investigated in this study. I focused on the NL2Flow domain, which combined service composition with goal-oriented dialogue to generate task workflows \citep{chakraborti2022natural}. The objective was to develop a system capable of autonomously selecting and composing a sequence of actions, which are termed ‘custom' actions, representing available services, to achieve user goals (e.g., submitting an expense report) (Listing \ref{lst:nl2flowplanexample}). Each `custom' action was modeled as a parameterized function, and successful execution necessitated providing values for a defined set of input parameters. Critically, these parameter values were frequently unknown at the initiation of a workflow, representing information the system needed to acquire from the user. To address this, the system used a dialogue-based approach. When a parameter value was missing, the system generated an ‘ask’ action, a dialogue act specifically designed to elicit the necessary information. This process of identifying and requesting missing parameter values was analogous to slot-filling, a common technique in task-oriented dialogue systems, and ensured the system possessed the information required to progress. Successfully executing a `custom' action updated the system’s internal state, transitioning the corresponding parameter from unknown to known, and enabling the system to proceed with subsequent actions in the workflow.

\subsection{Parametric Data Generation}

A parametric workflow problem generator was developed to create an intermediate representation (IR) of the problem space. Implemented in Python, this generator allows for controlled variation in workflow complexity and characteristics. A dataset of 2,296 workflows was generated using a fixed random seed (20250429) to ensure reproducibility. Workflow generation was governed by six key parameters (Table \ref{tab:generatorconf}): the number of `custom' actions (representing available operations), `custom' action arity (the number of input/output parameters defining each action’s interface), the number of goal actions (specifying the desired outcome), the proportion of slot-fillable variables (variables requiring values during planning), the complexity of variable mapping, and the proportion of coupled actions (actions where outputs serve as inputs to subsequent actions, creating dependencies). I aimed to generate ten unique samples for each valid parameter combination. `custom' actions were defined as callable components with specified input and output parameter sets, invoked when all required input parameters are known, and producing corresponding output parameters. These actions represent operations commonly found in agentic frameworks, such as agents and tools.

\begin{table}[H]
    \centering
    \caption{Workflow data generator configuration}
\begin{tabular}{ cc } 
  Parameters & Values \\ \hline
 `custom' actions & [2, 3, 4, 5] \\
 Arity & [1, 2, 3] \\ 
 Coupling & [0, 0.375, 0.626, 1.0] \\ 
 Slot-fillable variables & [0.375, 0.626, 1.0] \\
 Mapping & [0] \\ 
 Goals & [1, 2] \\ 
\end{tabular}
\label{tab:generatorconf}
\end{table}

To construct a workflow problem, 30 variables were generated and designated as either slot-fillable or initially known. `custom' actions were then created and connected to establish dependencies, with action outputs serving as inputs for subsequent actions. Variables not directly used within any action were assigned pre-defined values. A `custom' action is executable only when all required input parameter values are known; upon execution, the values of its output parameters become known. Goal actions were randomly selected from the `custom' action set. To focus on core workflow generation challenges, we excluded variable mapping and initial variable values from this study. The resulting intermediate representation (Python object) was compiled into PDDL strings using NL2Flow, enabling symbolic verification of large language model plans. Each generated problem instance was assigned a unique hash for identification and deduplication.

The parametric workflow problem generator constructed an intermediate representation (Listing \ref{lst:intermediaterepresentation} in Appendix), implemented as a Python object, to define the problem space. This representation comprises five key components: 

\begin{itemize}
    \item $available\_agents$ represent the available `custom' actions
    \item $actuator\_signature$ specifies the input and output parameters for each agent.
    \item $mappings$ enable the transfer of values between variables.
    \item $available\_data$ define the known values of variables at the workflow's outset.
    \item $goal\_agent\_ids$ are the ids of the desired actions to be achieved.
\end{itemize}

From this intermediate representation, I generated natural language prompts and a formal PDDL representation of the workflow problem. See Appendix, Prompt Generation and PDDL Compilation for details on prompt generation and PDDL compilation, respectively. Language model responses were parsed according to the procedure outlined in Appendix, Language Model Response Parsing.

\subsection{Plan Evaluation}

NL2Flow evaluated plans generated by LLMs using the Planning Domain Definition Language (PDDL). LLM-generated plans were extracted from text and converted to PDDL via NL2Flow, using an intermediate representation from the data generator. Details of this PDDL compilation process are provided in the PDDL Compilation subsection of the Appendix. The resulting PDDL formulations were then assessed with a symbolic planner. For clarity, a sound plan is defined as one that is executable; a valid plan is executable and achieves the goal state; and an optimal plan is executable, reaches the goal state, and does so with minimal cost. When a planning problem lacks a feasible solution, any proposed plan is invalid. Furthermore, identifying an optimal plan requires the existence of at least one feasible solution.

To assess the soundness of generated plans, the pddl compilation omits goal achievement. NL2Flow's approach distinguishes between soundness and validity. For soundness checking, the planning problem was compiled into PDDL, but instead of including the original goals as requirements, the actions reported as executed within the LLM's response were incorporated. This approach tests the LLM's claimed actions actually executable. Validity was checked by compiling with both original goals and actions in LLM's plan as requirements; failure to achieve goals indicates a non-valid plan where a planner finds a longer optimal plan. In this low-difficulty workflow problem setting, a plan was invalid, also, if the system prompted the user for a non-slot-fillable variable, or if a `custom' action was invoked with unknown input parameter values. I investigated the use of VAL \citep{howey2004val} for plan validation; however, VAL did not support the PDDL language features necessary to represent the complexities introduced by NL2Flow's PDDL compilation process, precluding its use in this evaluation.

To quantify plan optimality, the length of the LLM-generated plan was compared to that of optimal plans derived using the k$\ast$ planner \citep{lee-et-al-socs2023}. The planning problem was formulated in PDDL using only the original goals as requirements and use k$\ast$ to obtain a set of optimal plans. Following a successful validity check, an LLM-generated plan was considered sub-optimal if its length exceeded the length of the shortest k$\ast$-derived optimal plan. This comparison provides a quantitative measure of how efficiently the LLM solves the planning problem relative to a state-of-the-art planner.

Following normalization of the explanatory variables (Tables \ref{tab:lrnoplanlong}, \ref{tab:lrnoplanshort}, \ref{tab:lrplanvaliditylong}, \ref{tab:lrplanvalidityshort}, \ref{tab:lrplanoptimallong}, and \ref{tab:lrplanoptimalshort} in Appendix), I used logistic regression, implemented in the Python statsmodels package, to assess the contribution of these variables to different aspects of LLM workflow generation capacity.

\section{Results}
\subsection{Planning Task}
\subsubsection{Plan Length}

The length of LLM-generated plans exhibited considerable variation, ranging from 0 to 32 steps. However, the range of optimal plan lengths (as determined by K$\ast$), was narrower, spanning only 0 to 8 steps. A plan length of 0 indicates the model determined the goal was unattainable within its problem-solving capacity. Optimal plans of length one were not feasible due to the minimum arity of all actions being one; each action requires at least one input, necessitating a plan of at least two actions. While the majority of plans generated across all models were relatively concise (under 10 steps), several – DeepSeek-V3, Granite-3.3-8B-instruct, Llama-3.1-8B-instruct, Llama-3.3-70B-instruct, and Qwen2.5-72B-instruct – occasionally generated significantly longer plans. The longest observed plan (32 steps) was produced by Llama-3.1-8B-instruct using the verbose prompt style. Despite this outlier, extended plans (length$>$10) remained infrequent, comprising less than 1\% of all generated plans. In particular, CodeLlama-34B-instruct-hf and DeepSeek-Coder-33B-instruct consistently failed to produce plans conforming to the expected output format, preventing automated evaluation.

Kolmogorov-Smirnov tests indicated that prompt style did not significantly alter plan length distributions for DeepSeek-V3 and Llama-3.1-405b-instruct-fp8, but did significantly affect them for other models (Figures \ref{fig:planlengthverbose} and \ref{fig:planlengthconcise} in Appendix). Mann-Whitney U tests further showed that prompt style did not significantly influence the median plan length for DeepSeek-V3, Llama-3.1-405b-instruct-fp8, and Llama-3.3-70b-instruct, while it did impact the median plan length for the remaining models.

\subsubsection{Sound, Valid, and Optimal Planning}

\begin{figure}
\centering
  \includegraphics[scale=0.31]{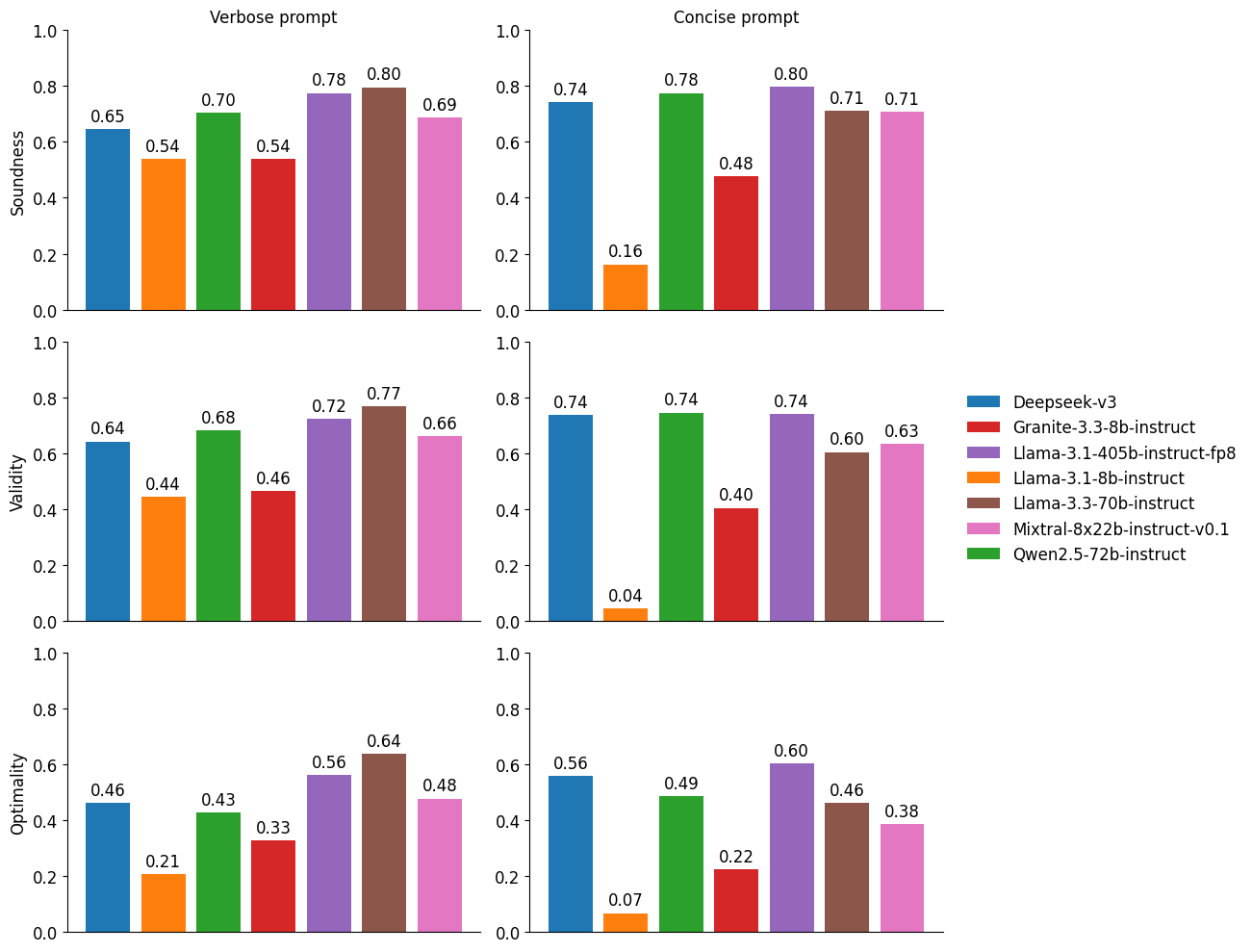}
  \caption{Comparison of LLM performance in generating sound, valid, and optimal plans using verbose and concise prompt styles}
  \label{fig:planall}
\end{figure}

Overall, Llama-3.3-70B-instruct, when paired with the verbose prompt style, demonstrated the strongest planning performance (Figure \ref{fig:planall}). This model achieved the highest rates of generating sound plans (80\%), valid plans (77\%), and optimal plans (64\%) among all tested models. Using the concise prompt style resulted in a noticeable decrease in performance, with rates dropping to 71\%, 60\%, and 46\% respectively. A similar, though more pronounced, trend was observed for Llama-3.1-8B-instruct. 

\subsubsection{No Plan}

\begin{figure}
\centering
  \includegraphics[scale=0.31]{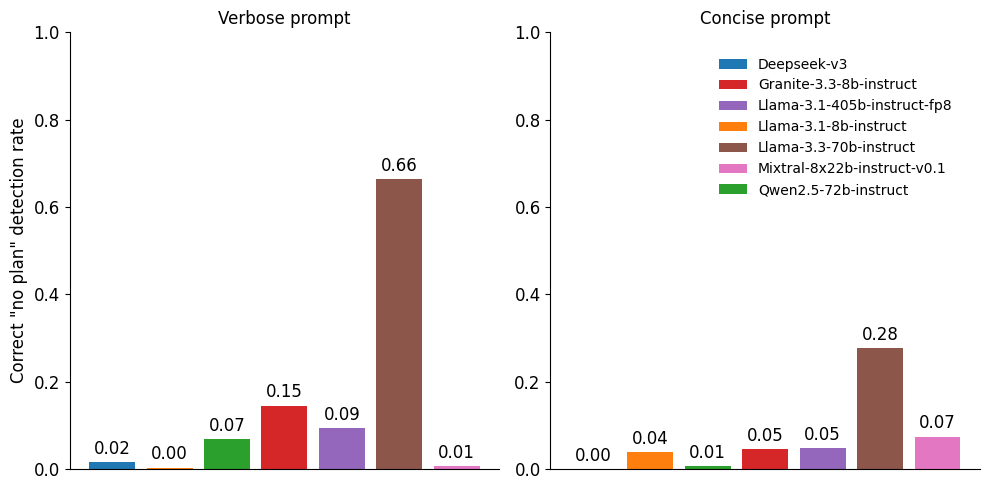}
  \caption{Comparison of large language model performance (accuracy) in detecting 'no plan' scenarios, evaluated with verbose and concise prompt styles}
  \label{fig:plannoplan}
\end{figure}

A key limitation observed across all models was the difficulty in correctly identifying unsolvable problems (Figure \ref{fig:plannoplan}) – those for which no plan exists. The superior performance of Llama-3.3-70B-instruct in generating sound, valid plans was largely attributable to its ability to recognize these ‘no plan’ scenarios. Analysis of explanatory variables – including the number of `custom' actions, arity, coupling, the proportion of slot-fillable variables, and the number of goals – showed that factors influencing the LLM's failure to correctly identify 'no plan' scenarios varied depending on the model and the prompt style (Tables \ref{tab:lrnoplanlong} and \ref{tab:lrnoplanshort} in Appendix). However, no single variable consistently predicted failure across all models or prompt styles.

\subsubsection{Feasible Plan}

\begin{figure}
\centering
  \includegraphics[scale=0.31]{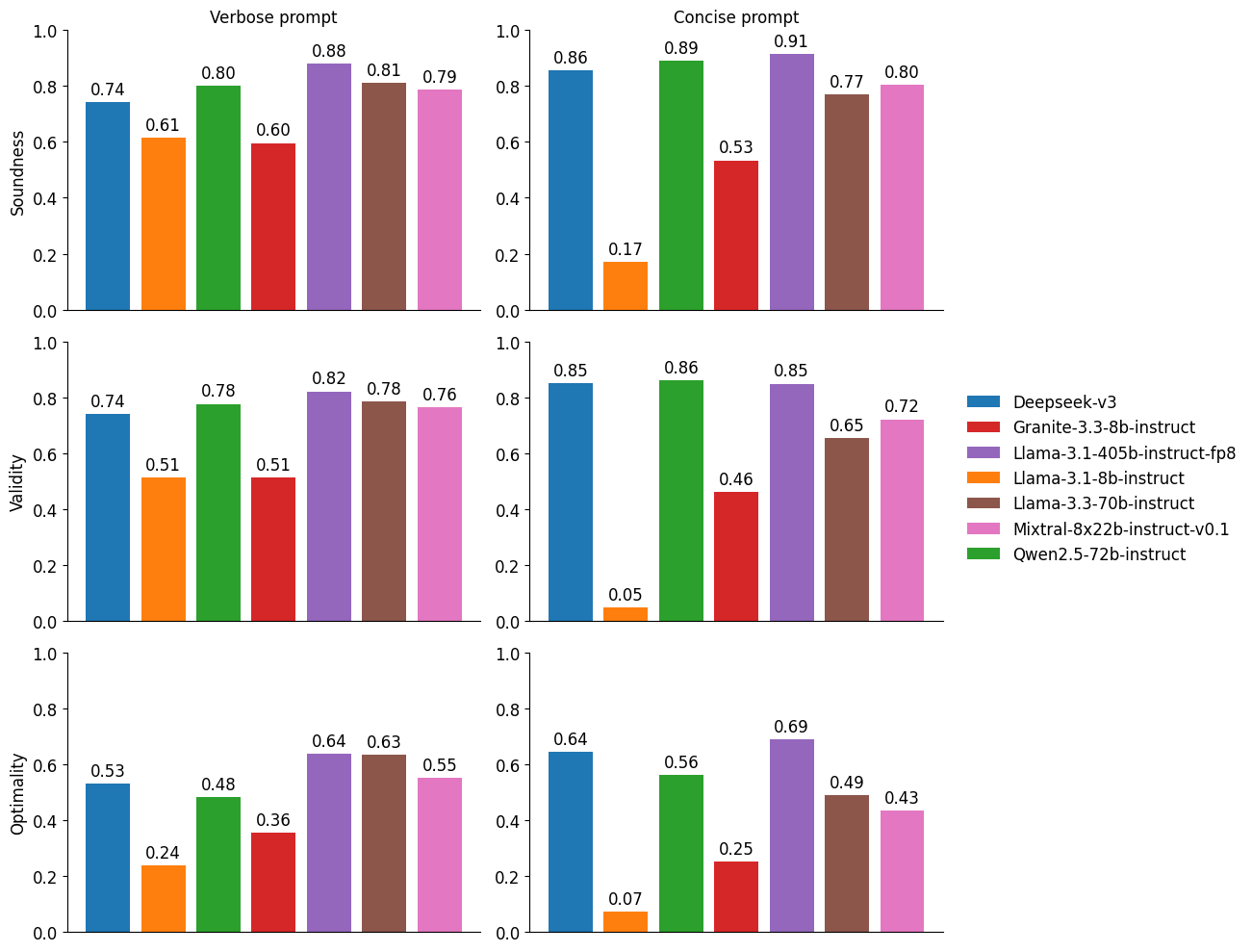}
  \caption{Performance comparison of large language models on planning tasks with feasible plans}
  \label{fig:planplan}
\end{figure}

When restricted to problems with feasible plans, Llama-3.1-405B-instruct-fp8 demonstrated the strongest performance, achieving the high rates of generating sound (91\%), valid (85\%), and optimal plans (69\%) using the concise prompt style (Figure \ref{fig:planplan}).  I further investigated the influence of problem characteristics – including optimal plan length, number of `custom' actions, arity, coupling, proportion of slot-fillable variables, and number of goals – on the generation of valid plans. The relationship between these explanatory variables and plan quality (soundness, validity, and optimality) varied considerably across language models.

Across most models, optimal plan length was the strongest predictor of valid plan generation (Tables \ref{tab:lrplanvaliditylong} and \ref{tab:lrplanvalidityshort} in Appendix). However, this relationship was not consistent.  With the verbose prompt style, logistic regression with L1 regularization ($\alpha = 1$) failed to identify any significant predictors of valid plan generation for Granite-3.3-8B-instruct and Llama-3.1-8B-instruct. Likewise, when using the concise prompt, optimal plan length was not a significant predictor for Llama-3.3-70B-instruct, Mixtral-8x22B-instruct-v0.1, and Qwen2.5-72B-instruct. These results indicate that the influence of problem difficulty on valid workflow generation is model-dependent.

Regarding optimal plan generation, coupling consistently explained a significant portion of the variance across all models when the verbose prompt was used (Table \ref{tab:lrplanoptimallong} in Appendix). Meanwhile, this trend was observed only in DeekSeek-V3 and Llama-3.3-70B-instruct when using the concise prompt style (Table \ref{tab:lrplanoptimalshort} in Appendix). These results suggest that the impact of problem characteristics on optimal plan generation is both model- and prompt-dependent.

\subsection{Natural Language to JSON Translation}

\begin{figure}
\centering
  \includegraphics[scale=0.38]{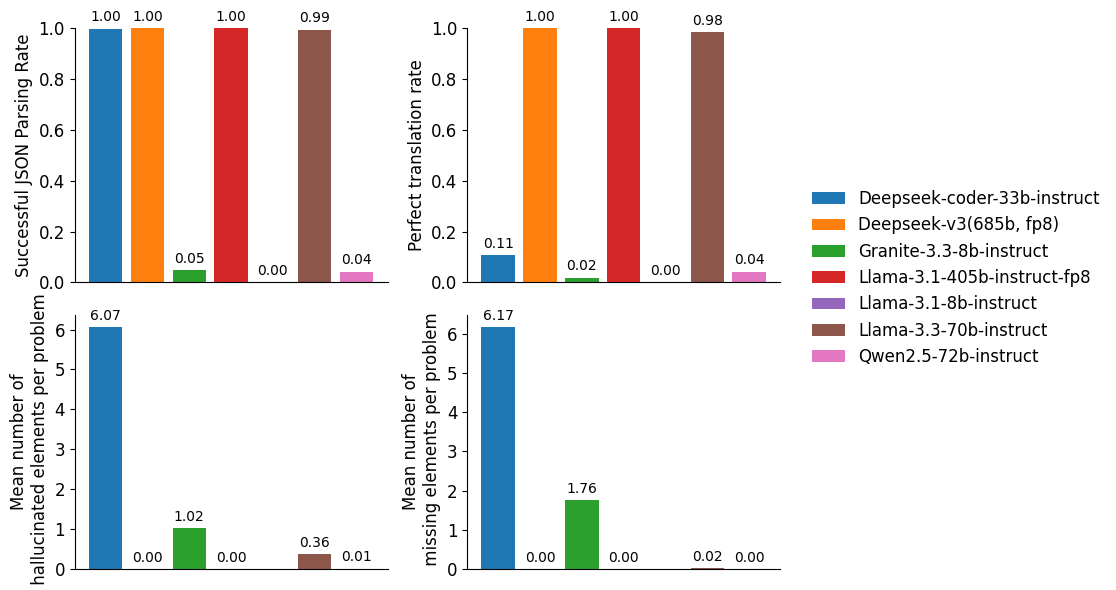}
  \caption{LLM-based natural language to JSON translation for planning}
  \label{fig:translationrates}
\end{figure}

DeekSeek-V3 and Llama-3.1-405B-instruct-FP8 consistently generated valid JSON strings (100\%) (Figure \ref{fig:translationrates}). Specifically, all JSON outputs from these models included all required elements and contained no hallucinated information. In contrast, DeepSeek-coder-33B-instruct produced valid JSON reliably, exhibiting a mean of over 6 hallucinated elements and over 6 missing elements. The models Granite-3.3-8B-instruct, Llama-3.1-8B-instruct, and Qwen2.5-72B-instruct failed to generate valid JSON in over 90\% of attempts.

\section{Related Work}

Current evaluation of reasoning capacity often focuses on final answer accuracy for benchmarks like MATH500 \citep{lightman2023let}, AIME24, GSM8K \citep{cobbe2021training}, and MATH. However, this metric overlooks the validity of the reasoning 'process' itself; a correct answer can be achieved through flawed reasoning, effectively rewarding a successful, yet invalid, plan. Furthermore, relying solely on final answer correctness leaves the quality and structure of the reasoning process opaque. Similarly, BIRD \citep{li2023can}, a natural language to SQL translation benchmark, primarily relies on final answer accuracy, despite the inherent planning aspects of SQL query construction alongside elements like intent detection. GSM-Symbolic \citep{mirzadeh2024gsm} attempts to address this by generating mathematical problems from symbolic templates, allowing for manual verification of solutions. ReasonAgain \citep{yu2024reasonagain} uses GPT-4o to generate Python programs for GSM8K and MATH, validating reasoning by executing the generated code.

Beyond these benchmarks, several studies use “puzzle” datasets \citep{shojaee2025illusion}, which include Tower of Hanoi, river crossing, blocks world, and checker jumping, to assess LLM reasoning. These puzzles offer the advantage of verifiable reasoning steps through simulation and can be scaled to challenge LLMs as they saturate. However, the applicability of these abstract puzzles to real-world scenarios remains limited.

A growing body of work focuses on evaluating and training LLMs for workflow automation and reasoning. Several datasets have been proposed to assess LLM planning capabilities, differing in their generation methods (static vs. dynamically generated) and evaluation approaches (manual vs. programmatic). This section details relevant work, highlighting these key distinctions.

\subsection{Static Datasets \& Primarily Manual/LLM-as-Judge Evaluation}

FlowBench \citep{xiao2024flowbench} presents a static dataset of 51 scenarios across 6 domains, focusing on multi-turn interactions. Evaluation relied heavily on human verification by comparing expected and observed traces, identifying a need for improved LLM planning. Similarly, HuggingGPT \citep{shen2023hugginggpt} uses a static, human-annotated dataset augmented with GPT-4 generated pseudo-labels, and uses LLM-as-a-judge for graph tasks. TaskLAMA \citep{yuan2024tasklama} also provides a static dataset, created through human annotation, for structured complex task decomposition. ISG-BENCH \citep{chen2024interleaved} focuses on interleaved text and image generation, using a static dataset and relying on MLLM-as-a-judge for scoring alignment, coherence, and helpfulness, grounded in human-annotated truth. WikiPlan and RecipePlan contribute static text-image pairs for task completion. These datasets, while valuable, often require significant human effort for creation and/or evaluation \citep{lu2023multimodal}.

\subsection{Static Datasets \& Programmatic Evaluation}

TaskBench \citep{shen2024taskbench} divides task automation into decomposition, tool selection, and parameter prediction, leveraging a static dataset of 17,331 samples generated with GPT-4. The authors used programmatic metrics (F1, Normalized Edit Distance, Rouge) for evaluation, with some partial manual inspection. This approach offers scalability due to a large sample size but can be limited by the quality and coverage of the initial dataset. Open Grounded Planning \citep{planningopen} also aims to generate executable plans, providing LLMs with an extensive action library and leveraging existing datasets. Evaluation combines programmatic metrics, such as executability and pass rate, with LLM-as-a-judge for assessing completeness, feasibility, and relevance.

\subsection{Algorithmically Generated Datasets \& Programmatic Evaluation}

WORFBENCH \citep{qiao2024benchmarking} represents a step towards automated evaluation through its data generator for graph workflow structures. LLM responses are evaluated via subgraph matching. However, there is a limitation: the need to define the entire graph upfront, hindering scalability – a limitation that search-based methods, like my own, avoid by not requiring complete graph definition for data generation and evaluation.

My work extends existing research by presenting a fully automated pipeline for data generation and evaluation, overcoming the limitations of static datasets and enabling scalable, programmatic assessment of LLM workflow generation. However, the effectiveness of this approach is ultimately bounded by the capacity of the symbolic planner to evaluate workflow optimality.

\section{Conclusion}

This work demonstrates that current open-source, instruct-tuned LLMs can generate effective plans directly from natural language instructions, outperforming previous LLM planning studies that relied on formal PDDL representations \citep{silver2022pddl, valmeekam2023planning, kokel2025acpbench}. This difference likely arises from the nature of the reasoning tasks: our study focused on workflow creation, where generated actions inherently involve fixed input and output variables, contrasting with the flexible variable binding common in domains like the Blocks World. Importantly, the complexity of workflow creation is comparable to that of domains used in prior LLM planning research. The observed pattern reinforces the hypothesis that LLM planning capabilities are domain-specific, potentially rooted in their statistical reliance on token prediction and the associated sensitivity to the compositional complexity of input tokens \citep{zhao2025chain}.

Regression analysis confirmed that the influence of problem characteristics on the goodness of plan is dependent on both the language model and the prompting strategy employed. This highlights the importance of model- and prompt-specific analysis, cautioning against broad generalizations across LLMs.  My study demonstrated that translating natural language into an intermediate JSON workflow representation significantly improved performance compared to direct plan generation (Figure \ref{fig:overview}). This suggests that prematurely replacing established neuro-symbolic planning systems with direct language-to-action reasoning may lead to a decline in plan quality. While decomposing planning into multiple steps can introduce complexity, simplifying the system by directly reasoning from language to action can, in some applications, prioritize reduced complexity over plan quality.  However, as LLM reasoning capabilities advance and scale to more complex problems, the limitations and error sources within AI planning systems are likely to evolve.

NL2Flow extends beyond simple LLM assessment by providing a framework for evaluating the planning capabilities of novel agentic architectures. These architectures vary in agent and tool configurations and operate with randomly assigned goals, expressed as LLM-generated utterances based on system-defined agent and tool descriptions (including those adhering to the Model Context Protocol (MCP)). By translating these goals and configurations into PDDL, NL2Flow enables the evaluation of agentic system trajectories using a standardized planning framework, allowing for assessment of an agent’s ability to select and compose tools based on custom, symbolically-verifiable workflows. Given the prevalence of chain-of-thought, tree-of-thought, and self-consistency prompting methods in agentic systems for workflow creation, verifying the soundness of this semantic approach is crucial for improving system reliability.

Future work will focus on expanding this automated pipeline to generate datasets exhibiting greater combinatorial complexity and incorporating diverse constraints. Investigating the interplay between prompting strategies, model architecture, and dataset characteristics will be crucial for understanding how to best use LLMs for complex workflow automation. Ultimately, a systematic analysis of LLM limitations across a spectrum of reasoning tasks will be vital for guiding future development and realizing the full potential of these models.

\section{Acknowledgements}

The author gratefully acknowledges the close collaboration with Dr. Tathagata Chakraborti, whose guidance was instrumental in the design and implementation of NL2FLOW, particularly in the development of its core planning algorithm.  The author also thanks Dr. Junkyu Lee at IBM Research for providing valuable feedback on an early draft of this manuscript.

\bibliographystyle{plain} 
\bibliography{neurips2025}

\section*{Appendix}
\renewcommand{\thefigure}{A\arabic{figure}} 
\renewcommand{\thetable}{A\arabic{table}}
\renewcommand{\thelstlisting}{A\arabic{lstlisting}}
\setcounter{figure}{0}  
\setcounter{lstlisting}{0}  
\setcounter{table}{0}  

\subsection*{Intermediate Representation Example}

\begin{center}
\begin{lstlisting}[basicstyle=\small, frame=tb, breaklines=true, numbers=none, label={lst:intermediaterepresentation}, caption={Intermediate representation from data generator. An example of the intermediate representation is provided in the supplementary material as `intermediate\_representation.json`.}]
{
    "available_agents": [
        {
            "agent_id": "a__2",
            "actuator_signature": {
                "in_sig_full": [
                    {
                        "name": "v__7",
                        "data_type": null,
                        "mappable": false,
                        "slot_fillable": true
                    }
                ],
                "out_sig_full": [
                    {
                        "name": "v__9",
                        "data_type": null,
                        "mappable": false,
                        "slot_fillable": true
                    }
                ]
            }
        },
        ...
    ],
    "goal_agent_ids": [
        "a__1",
        "a__0"
    ],
    "mappings": [],
    "available_data": [
        [
            "v__10",
            null
        ],
        ...
    ]
}
\end{lstlisting}
\end{center}

\subsection*{Prompt Generation}

This study examined the capacity of LLMs to generate executable plans to achieve specified goals and to translate natural language problem descriptions into JSON schemas. I adopted a few-shot prompting approach, constructing prompt styles for both tasks and providing six example demonstrations. Crucially, each example within the prompt presented a natural language description of the plan execution before the corresponding plan expressed in the target format (Listing \ref{lst:responseexample}). This was intended to guide the LLM by demonstrating the logical connection between action sequences and their executable representations, thereby increasing the likelihood of generating valid and goal-directed plans.

\begin{center}
\begin{lstlisting}[basicstyle=\small, frame=tb, breaklines=true, numbers=none, label={lst:responseexample}, caption={Planning response example}]
PLAN EXPLANATION
0. Execute action a__2. This will result in acquiring v__1. v__1 is required later by action A.
1. Execute action a__1 with aa as input. This will result in acquiring v__2. v__2 is required later by action a__3. Since executing A was a goal of this plan, return the results of A(v__1) to the user.
2. Execute action a__3 with v__2 as input. Since executing a__3 was a goal of this plan, return the results of a__3(v__2) to the user.
PLAN
[0] v__1 = a__2()
[1] v__2 = a__1(v__1)
[2] a__3(v__2)
\end{lstlisting}
\end{center}

For the planning task, I investigated the impact of prompt style, comparing `verbose' (Listing \ref{lst:longprompt}) and `concise' (Listing \ref{lst:shortprompt}) prompt styles. The `verbose' prompts (mean length 13,413 characters) included a detailed, natural language description of all relevant elements within the planning problem, aiming to provide the LLM with comprehensive contextual information. Conversely, `concise' prompts (mean length 9,674 characters) presented these elements as a concise, enumerated list, minimizing verbose text and focusing on key parameters. Listings \ref{lst:longpromptaction} and \ref{lst:shortpromptaction} show the description of `custom' action $a\_\_1$ in verbose and concise prompts, respectively. This variation was designed to assess the LLM’s sensitivity to the level of contextual detail provided in the prompt.

\begin{center}
\begin{lstlisting}[basicstyle=\small, frame=tb, breaklines=true, numbers=none, label={lst:longpromptaction}, caption={`custom' action in verbose prompt}]
Action a__1 has input parameters v__11, v__12, and v__13 and it outputs variables v__15, v__16, and v__18. To execute action a__1, variables v__11, v__12, and v__13 must be known. After executing action a__1, variables v__15, v__16, and v__18 are known.
\end{lstlisting}
\end{center}

\begin{center}
\begin{lstlisting}[basicstyle=\small, frame=tb, breaklines=true, numbers=none, label={lst:shortpromptaction}, caption={`custom' action in concise prompt}]
name: a__1
inputs: v__11, v__12, v__13
outputs: v__15, v__16, v__18
\end{lstlisting}
\end{center}

For translation, a single prompt style (Listing \ref{lst:translationprompt}) was used, designed to elicit accurate and fluent translations. The complete prompt styles used for each task – verbose\_planning\_prompt.txt, concise\_planning\_prompt.txt, and translation\_prompt.txt – are included in the supplementary material to facilitate reproducibility, enable independent verification of my results, and encourage further research in this area. Each prompt style followed a consistent structure comprising three sections: Context, Instruction, and Examples (Listings \ref{lst:longprompt} and \ref{lst:shortprompt}). The Context section provided a programmatically generated natural language description of the problem. The Instruction section detailed the desired planning process, specifying preferences for plan structure, the required output format, and guidance for handling unsolvable problems. To facilitate parsing and execution, plans were to be expressed in a Python-like syntax (e.g., outputs = \textlangle action name\textrangle(\textlangle parameters\textrangle)). I standardized action names as a\_\_\textlangle number\textrangle and variable names as v\_\_\textlangle number\textrangle  to promote consistency and reduce ambiguity.

\begin{center}
\begin{lstlisting}[basicstyle=\small, frame=tb, breaklines=true, numbers=none, label={lst:longprompt}, caption={Verbose planning prompt}]
CONTEXT
The system has actions a__0, a__1, and a__2.
Action a__1 has input parameters v__13, v__12, and v__11 and it outputs variables v__15, v__16, and v__18. To execute action a__1, variables v__13, v__12, and v__11 must be known. After executing action a__1, variables v__15, v__16, and v__18 are known.
Action a__2 has ...

INSTRUCTION
Develop a concise plan to achieve the goals. The shorter plan is preferable. A plan consists of individual actions, each represented by its outputs and an action name followed by its parameters in parentheses (e.g., outputs = <action-name>(<parameters>). When it becomes impossible to achieve the objectives, the appropriate course of action should be considered as `no plan'

Here are some examples.
Example #1 ...

Now output your plan explanation and plan.
{plan_explanation_and_plan}
\end{lstlisting}
\end{center}

\begin{center}
\begin{lstlisting}[basicstyle=\small, frame=tb, breaklines=true, numbers=none, label={lst:shortprompt}, caption={Concise planning prompt}]
CONTEXT
The system has actions a__0, a__1, and a__2. To execute an action the values of their inputs must be known.

actions:

name: a__1
inputs: v__11, v__12, v__13
outputs: v__15, v__18, v__16

name: a__2
inputs: ...

INSTRUCTION
Develop a concise plan to achieve the goals. The shorter plan is preferable. A plan consists of individual actions, each represented by its outputs and an action name followed by its parameters in parentheses (e.g., outputs = <action-name>(<parameters>). When it becomes impossible to achieve the objectives, the appropriate course of action should be considered as `no plan'

Here are some examples.
Example #1 ...

Now output your plan explanation and plan.
{plan_explanation_and_plan}
\end{lstlisting}
\end{center}

This study evaluated nine LLMs across translation and planning tasks: CodeLlamma-34B-instruct, DeepSeek-coder-33B-instruct, DeepSeek-V3(685B, fp8), Granite-3.3-8B-instruct, Llama-3.1-8B-Instruct, Llama-3.1-405B-instruct-fp8, Llama-3.3-70B-instruct, Mixtral-8x22B-instruct-v0.1, and Qwen2.5-72B-instruct.

\begin{center}
\begin{lstlisting}[basicstyle=\small, frame=tb, breaklines=true, numbers=none, label={lst:translationprompt}, caption={Translation prompt}]
CONTEXT
The system has actions a__0, a__1, and a__2.
Action a__1 has input parameters v__13, v__12, and v__11 and it outputs variables v__15, v__16, and v__18. To execute action a__1, variables v__13, v__12, and v__11 must be known. After executing action a__1, variables v__15, v__16, and v__18 are known.
Action a__2 has ...

INSTRUCTION
Write a json to describe the context.


Here are some examples.
Example #1 ...

Now output your json.
{json}
\end{lstlisting}
\end{center}

\subsection*{Language Model Response Parsing}

To establish a lower bound on large language model performance, I used minimal parsing techniques for both translation and planning tasks, prioritizing simplicity and avoiding potentially biasing post-processing steps.

For the translation task, model outputs were structured as JSON strings representing a planning problem. These strings were then parsed using PyDantic, a data validation and settings management library, to ensure adherence to a predefined schema. This structured format enabled automated evaluation metrics and facilitated downstream analysis. The initial intermediate representation schema used nested structures to represent `custom' actions; however, this complexity proved unnecessary, as the same information could be represented with a simplified schema. To evaluate the capacity of LLMs to translate natural language descriptions of workflow problems into JSON, I used a streamlined schema (Listing \ref{lst:translationobject}) defined as follows:

\begin{itemize}
    \item $actions$ represents `custom' actions.
    \item $mappings$ corresponds to the `mappings' present in the intermediate representation.
    \item $available\_data$ is a list of variables with known values at the workflow's initiation.
    \item $askable\_parameters$ is a list of slot-fillable variables for user input.
    \item $unaskable\_parameters$ is a list of variables for which the system is prohibited from prompting the user.
    \item $goal\_action\_ids$ corresponds to the `goal\_action\_ids' in the intermediate representation.
\end{itemize}

An example of this schema, used for evaluating LLM translation capacity, is provided as `translation\_object.json' in the supplementary material.

For the planning task, each action within a generated plan was extracted and stored as a string within a Python list. This list was subsequently used as input to the NL2Flow debugger, a symbolic planning validation tool that assesses the logical consistency and feasibility of the generated plan.

\begin{center}
\begin{lstlisting}[basicstyle=\small, frame=tb, breaklines=true, numbers=none, label={lst:translationobject}, caption={JSON schema used for evaluating LLM's translation capacity}]
{
    "actions": [
        {
            "id": "a__1",
            "input": [
                "v__1",
                "v__2"
            ],
            "output": [
                "v__3"
            ]
        },
        ...
    ],
    "mappings": [],
    "available_data": [
        "v__5",
        ...
    ],
    "askable_parameters": [
        "v__1",
        ...
    ],
    "unaskable_parameters": [
        "v__4",
        ...
    ],
    "goal_action_ids": [
        "a__3"
    ]
}
\end{lstlisting}
\end{center}

\subsection*{AI Planning}

AI planning addresses the problem of finding a sequence of actions that transforms an initial state into a goal state. Formally, a planning problem is defined as $P = \langle P, F, D, I, G \rangle$, where:

\begin{itemize}
    \item $P$ is a finite set of predicates describing properties of the world.
    \item $F$ is a finite set of numeric variables.
    \item $D$ is the domain model, comprising a set of action schemas.
    \item $I$ is the initial state, describing the world at the start of planning.
    \item $G$ is the goal state, specifying the desired outcome.
\end{itemize}

I adopt the closed-world assumption, meaning any fact not explicitly stated in the initial state is considered false.  An action $a$ is defined by its preconditions and effects: $a = \langle pre(a), eff(a) \rangle$. Preconditions, $pre(a)$, define the conditions that must hold in a state for the action to be applicable. Effects, $eff(a)$, specify how the state changes upon execution of the action.

A state $s$ is a conjunction of predicates and numeric variable assignments: $s = \langle P_s, V_s \rangle$, where $P_s \subseteq P$ represents the set of true predicates, and $V_s \in \mathbb{Q}^{|F|}$ represents the assignment of rational values from $\mathbb{Q}$ to each numeric variable in $F$.

Conditions are expressed as a combination of predicate literals and numeric constraints.  A numeric constraint is a relation between arithmetic expressions: $\langle exp_1, \triangleright, exp_2 \rangle$, where $\triangleright \in \{<, \leq, >, \geq, =\}$. Arithmetic expressions ($exp$) are built from numeric variables in $F$, rational constants in $\mathbb{Q}$, and the operators +, and -.

An effect $eff(a)$ is a triple $\langle P^+_{eff}, P^-_{eff}, U_{eff} \rangle$, where:

\begin{itemize}
    \item $P^+_{eff} \subseteq P$ is the set of predicates that become true.
    \item $P^-_{eff} \subseteq P$ is the set of predicates that become false.
    \item $U_{eff}$ is a set of numeric updates, each of the form $\langle f, :=/+=/-=, exp \rangle$, where $f \in F$ is the updated variable and $exp$ is the new value or change in value.
\end{itemize}

Applying an action $a$ in state $s$ is possible if all predicates in $pre(a)$ are present in $P_s$ and all numeric constraints in $pre(a)$ hold true when evaluated using $V_s$. The resulting state $s' = \langle P'_s, V'_s \rangle$, is computed as follows: $P'_s = (P_s \setminus P^-_{eff}) \cup P^+_{eff}$, and each variable $f \in F$ is updated in $V'_s$ according to the operator and expression specified in $U_{eff}$.

A plan $\pi$ is a sequence of actions. The trajectory of a plan, starting from initial state $I$, is the sequence of states $s_0, s_1, \dots$

\subsection*{PDDL Compilation}

This study used NL2Flow to translate an intermediate representation of a workflow creation problem into PDDL. The PDDL formulation within NL2Flow enabled cost-based planning to identify efficient workflows (plans). While NL2Flow was designed to represent a broad range of workflow automation concepts, this study focused on generating PDDL for simple, low-difficulty workflows. For instance, the mapping capacity feature, which allowed variable values to be shared, was excluded from the current implementation.

Planning problems in this work were formalized using the Planning Domain Definition Language (PDDL). PDDL separated each problem’s description into a domain model, which defined the possible actions and their effects, and a problem instance, specifying the initial state and goal conditions. Example domain and problem files used in this study (domain.pddl and problem.pddl, respectively) were provided as supplementary material. A domain model included declarations of predicates, functions, and actions, while a problem instance defined the initial state through a set of grounded predicates and function assignments. Any fact not explicitly stated as true in the initial state was considered false. Grounding involved replacing variables in predicates and functions with concrete values.

\subsubsection*{Types}

The domain defines several types:

\begin{itemize}
    \item $generic$: Represents the objects being processed – the entities whose slots need filling.
    \item $operator$: Represents actions the agent can take.
    \item $has-done-state$:  Represents temporal states (past, present, future) to track whether an action has been performed.
    \item $constraint-status$: Represents certainty levels (True, False) likely used in constraints or conditions.
    \item $datum-state$: Represents the certainty of information about an object (certain, uncertain, unknown).
    \item $label$:  (Currently unused) labeling objects or slots.
    \item $object$: A base type for extensibility.
\end{itemize}

\subsubsection*{Constants}

Many constants are defined for each type.  The `generic' constants ($v\_\_x$) represent specific instances of the objects being processed.  The $has{-}done{-}state$ constants track the temporal status of actions.  The $constraint{-}status$ and $datum{-}state$ constants define the possible values for these attributes. $operator$ constants ($a\_\_x$) represent specific, pre-defined actions.

\subsubsection*{Predicates}

The predicates define the relationships and properties within the domain:

\begin{itemize}
    \item $(has\_done\: ?x\: -\: operator\: ?x2\: -\: has{-}done{-}state)$: Indicates that an operator has been performed in a specific time state.
    \item $(been\_used\: ?x\: -\: generic)$:  Marks an object as having been processed.
    \item $(new\_item\: ?x\: -\: generic)$:  Indicates an object is newly introduced.
    \item $(known\: ?x\: -\: generic\: ?x2\: -\: datum{-}state)$:  Represents the agent's knowledge about an object's certainty. This is central to the planning process.
    \item $(not\_slotfillable\: ?x\: -\: generic)$: Indicates an object cannot have its slots filled (perhaps due to inherent properties).
    \item $(is\_mappable\: ?x\: -\: generic\: ?x2\: -\: generic)$ and $(not\_mappable\: ?x\: -\: generic\: ?x2\: -\: generic)$:  The ability to map one object to another.
    \item $(mapped\: ?x\: -\: generic)$: Indicates an object has been mapped.
    \item $(not\_usable\: ?x\: -\: generic)$: Indicates an object cannot be used.
    \item $(mapped\_to\: ?x\: -\: generic\: ?x2\: - generic)$:  Indicates an object is mapped to another object.
    \item $(free\:?x\:-\:generic)$: Indicates an object is available for processing.
    \item $(done\_goal\_pre)$ and $(done\_goal\_post)$:  Flags to indicate the goal has been achieved before and after certain actions.
\end{itemize}

\subsubsection*{Functions}

Functions are used to represent numeric state variables. They allow you to model quantities that change over time and are used in numeric effects of actions.

\begin{itemize}
    \item $(total{-}cost)$: A numeric function representing the cumulative cost of the plan.
    \item $(slot\_goodness\: ?x\: -\: generic)$: A numeric function that likely evaluates the \"quality\" or usefulness of filling a slot for a given object.
    \item $(affinity\: ?x\: -\: generic\: ?x2\: -\: generic)$: A numeric function that likely measures the similarity or relatedness between two objects.
\end{itemize}

\subsubsection*{Actions}

The domain defines two action types. `custom' actions are named according to the pattern $a\_\_x$.

\begin{itemize}
    \item $a\_\_x$:
        \begin{itemize}
            \item Precondition: Require that the action hasn't been done in the past and that specific objects ($v\_\_x$) are known with certainty.
            \item Effect: Mark the action as done in the present. Mark the object as used. Free up objects. Mark the newly freed object as known with certainty. Mark the newly freed object as not mapped. Increase the $total{-}cost$ by 10.
        \end{itemize}
\end{itemize}

\begin{center}
\begin{lstlisting}[basicstyle=\small, frame=tb, breaklines=true, numbers=none, label={lst:pddlactionexample}, caption={`custom' action example in PDDL}]
(:action a__2
    :parameters ()
    :precondition (and (not (has_done a__2 past)) (known v__7 certain))
    :effect (and
        (has_done a__2 present)
        (been_used v__7)
        (free v__9)
        (known v__9 certain)
        (not (mapped v__9))
        (increase (total-cost ) 10)
    )
)
\end{lstlisting}
\end{center}

For example, Listing \ref{lst:pddlactionexample} illustrate the structure of `custom' actions defined in the domain model. This action, $a\_\_2$, has no parameters. Its precondition requires that the action hasn’t been performed in the past ($not (has\_ done\: a\_\_2\: past)$) and that a variable $v\_\_7$ is known with certainty ($known\: v\_\_7\: certain$). The action's effect marks the action as done in the present ($has\_ done\: a\_\_2\: present$), indicates $v\_\_7$ has been used, makes $v\_\_9$ free, establishes $v\_\_9$ as known with certainty, indicates $v\_\_9$ is no longer mapped, and increases the $total\mbox{-}cost$ by 10.

The $ask$ action prompts the system to request information from the user if parameters are slot-fillable (Listing \ref{lst:pddlaskactionexample}).

\vspace{1.5em} 
\begin{center}
\begin{lstlisting}[basicstyle=\small, frame=tb, breaklines=true, numbers=none, label={lst:pddlaskactionexample}, caption={`ask' action example in PDDL}]
(:action ask
    :parameters (?x - generic)
    :precondition (and (not (known ?x certain)) (not (not_slotfillable ?x))
    )
    :effect (and
        (free ?x)
        (mapped_to ?x ?x)
        (known ?x certain)
        (not (not_usable ?x))
        (not (mapped ?x))
        (increase (total-cost ) (slot_goodness ?x))
    )
)
\end{lstlisting}
\end{center}

\begin{itemize}
    \item `ask':
        \begin{itemize}
            \item Takes a $generic$ object as a parameter.
            \item Precondition: Requires that the object is not currently known with certainty and is slotfillable ($not\: not\_slotfillable$).
            \item Effect: Frees the object. Maps the object to itself (a placeholder). Marks the object as known with certainty. Marks the object as usable (not $not\_usable$). Marks the object as not mapped. Increases the $total-cost$ by the $slot\_goodness$ of the object.
        \end{itemize}
\end{itemize}

\subsubsection*{Problem}

A problem describes a specific instance of a planning problem. It works in conjunction with a domain, which defines the general rules and actions applicable to a class of problems. A problem contains:

\begin{itemize}
    \item Initial state: The starting situation for the planner.  This is a description of what's true at the beginning. Predicates not explicitly defined as true in the initial state of a problem are considered false (closed-world assumption).
    \item Goal state: What the planner needs to achieve to solve the problem.  This is a condition that must be true when the plan is complete.
    \item Specifies objects: The specific objects that exist in this particular problem instance.
\end{itemize}

The initial condition of the problem, as described in `problem.pddl' in supplementary material, can be summarized as follows:

\begin{itemize}
    \item $slot\_goodness$ is defined for all generic objects ($v\_\_x$ and $new\_object\_generic\_0$).
    \item $total-cost$ is initialized to 0.0.
    \item For any generic object $v\_\_x$ used as an input parameter to a `custom' action, the predicate $(been_used\:v\_\_x)$ is true.
    \item The predicate $(new\_item\:new\_object\_generic\_0)$ is true.
    \item For all generic object, predicates $(mapped\_to\:?x\:?x)$ are true. In other words, All generic objects are self-mapped.
\end{itemize}

In this study, the initial state was defined by a set of variables ($v\_\_x$), each assigned a $slot\_goodness$ value of 1000000 if slot-fillable, and 150000 if not. These differing $slot\_goodness$ values represented a `soft' constraint, which was not used in this study because the precondition for the `ask' action prevented the system from requesting information for variables that were not slot-fillable. The $mapped\_to$ predicate establishes a self-mapping for each resource, essentially stating that each resource is initially assigned to itself. A $new\_item$ predicate indicates the creation of a new resource. The goal is to achieve conditions to complete a task or tasks ($and\:(has\_done\: a\_\_x\: present)(has\_done\: a\_\_x\: present)$) while minimizing the $total-cost$.

\subsection*{Regression Analysis Results}

\begin{center}

\begin{longtable}{p{3cm}p{1.5cm}p{1.5cm}p{1cm}p{1cm}p{1.5cm}p{1.5cm}}
\caption{Logistic regression analysis for no plan with verbose prompt style}
\label{tab:lrnoplanlong}
\small
\\ \hline
 DeepSeek-V3                 &                &         &           &         &          &          \\
Pseudo R-squ                & Log-Likelihood &         &           &         &          &          \\
-0.1845                     & -48.236        &         &           &         &          &          \\
                            & coef           & std err & z         & $P>|z|$ & {[}0.025 & 0.975{]} \\
`custom' actions                     & -20.9808       & 14.949  & -1.403    & 0.16    & -50.281  & 8.319    \\
arity                       & -19.1153       & 15.123  & -1.264    & 0.206   & -48.755  & 10.524   \\
coupling                    & 0              & nan     & nan       & nan     & nan      & nan      \\
slot-fillable variables     & 0              & nan     & nan       & nan     & nan      & nan      \\
goals                       & -2.3799        & 11.24   & -0.212    & 0.832   & -24.409  & 19.649   \\
\hline Granite-3.3-8B-instruct     &                &         &           &         &          &          \\
Pseudo R-squ                & Log-Likelihood &         &           &         &          &          \\
-0.0137                     & -146.9         &         &           &         &          &          \\
                            & coef           & std err & z         & $P>|z|$ & {[}0.025 & 0.975{]} \\
`custom' actions                     & -0.0534        & 10.769  & -0.005    & 0.996   & -21.161  & 21.054   \\
arity                       & -15.7939       & 11.914  & -1.326    & 0.185   & -39.144  & 7.556    \\
coupling                    & -6.1441        & 4.01    & -1.532    & 0.126   & -14.004  & 1.716    \\
slot-fillable variables     & -8.00E-17      & 11.144  & -7.18E-18 & 1       & -21.841  & 21.841   \\
goals                       & 0              & nan     & nan       & nan     & nan      & nan      \\
\hline Llama-3.1-405B-instruct-fp8 &                &         &           &         &          &          \\
Pseudo R-squ                & Log-Likelihood &         &           &         &          &          \\
-0.03309                    & -110.76        &         &           &         &          &          \\
                            & coef           & std err & z         & $P>|z|$ & {[}0.025 & 0.975{]} \\
`custom' actions                     & -14.7038       & 12.09   & -1.216    & 0.224   & -38.4    & 8.993    \\
arity                       & -11.7652       & 13.046  & -0.902    & 0.367   & -37.335  & 13.805   \\
coupling                    & 0              & nan     & nan       & nan     & nan      & nan      \\
slot-fillable variables     & -2.3115        & 13.035  & -0.177    & 0.859   & -27.86   & 23.237   \\
goals                       & 0              & nan     & nan       & nan     & nan      & nan      \\
\hline 
Llama-3.3-70B-instruct      &                &         &           &         &          &          \\
Pseudo R-squ                & Log-Likelihood &         &           &         &          &          \\
0.007233                    & -189.65        &         &           &         &          &          \\
                            & coef           & std err & z         & $P>|z|$ & {[}0.025 & 0.975{]} \\
`custom' actions                     & 11.4596        & 8.745   & 1.31      & 0.19    & -5.681   & 28.6     \\
arity                       & -7.33E-17      & 8.774   & -8.36E-18 & 1       & -17.196  & 17.196   \\
coupling                    & -2.0446        & 3.507   & -0.583    & 0.56    & -8.918   & 4.829    \\
slot-fillable variables     & 0              & nan     & nan       & nan     & nan      & nan      \\
goals                       & 0              & nan     & nan       & nan     & nan      & nan      \\
\hline Llama-3.1-8B-instruct       &                &         &           &         &          &          \\
Pseudo R-squ                & Log-Likelihood &         &           &         &          &          \\
-0.08881                    & -65.393        &         &           &         &          &          \\
                            & coef           & std err & z         & $P>|z|$ & {[}0.025 & 0.975{]} \\
`custom' actions                     & -21.3022       & 13.77   & -1.547    & 0.122   & -48.291  & 5.686    \\
arity                       & -17.1386       & 13.92   & -1.231    & 0.218   & -44.421  & 10.144   \\
coupling                    & 0              & nan     & nan       & nan     & nan      & nan      \\
slot-fillable variables     & 0              & nan     & nan       & nan     & nan      & nan      \\
goals                       & -8.56E-17      & 10.337  & -8.28E-18 & 1       & -20.261  & 20.261   \\
\hline Mixtral-8x22B-instruct-v0.1 &                &         &           &         &          &          \\
Pseudo R-squ                & Log-Likelihood &         &           &         &          &          \\
-0.05837                    & -57.104        &         &           &         &          &          \\
                            & coef           & std err & z         & $P>|z|$ & {[}0.025 & 0.975{]} \\
`custom' actions                     & -37.7334       & 5.594   & -6.746    & 0       & -48.697  & -26.77   \\
arity                       & 0              & nan     & nan       & nan     & nan      & nan      \\
coupling                    & -3.1613        & 5.535   & -0.571    & 0.568   & -14.009  & 7.687    \\
slot-fillable variables     & 0              & nan     & nan       & nan     & nan      & nan      \\
goals                       & 0              & nan     & nan       & nan     & nan      & nan      \\
\hline Qwen2.5-72B-instruct        &                &         &           &         &          &          \\
Pseudo R-squ                & Log-Likelihood &         &           &         &          &          \\
-0.01343                    & -92.919        &         &           &         &          &          \\
                            & coef           & std err & z         & $P>|z|$ & {[}0.025 & 0.975{]} \\
`custom' actions                     & -0.7135        & 12.438  & -0.057    & 0.954   & -25.091  & 23.664   \\
arity                       & -24.2876       & 12.676  & -1.916    & 0.055   & -49.132  & 0.557    \\
coupling                    & -6.8198        & 4.877   & -1.398    & 0.162   & -16.378  & 2.739    \\
slot-fillable variables     & 0              & nan     & nan       & nan     & nan      & nan      \\
goals                       & -1.7047        & 9.061   & -0.188    & 0.851   & -19.464  & 16.054 \\ \hline
\end{longtable}
\end{center}


\begin{center}
\begin{longtable}{p{3cm}p{1.5cm}p{1.5cm}p{1cm}p{1cm}p{1.5cm}p{1.5cm}}
\caption{Logistic regression analysis for no plan with concise prompt style}
\label{tab:lrnoplanshort} 
\small
\\
\hline 
DeepSeek-V3                 &                &         &           &         &          &          \\
Pseudo R-squ                & Log-Likelihood &         &           &         &          &          \\
inf                     & -18.971        &         &           &         &          &          \\
                            & coef           & std err & z         & $P>|z|$ & {[}0.025 & 0.975{]} \\
`custom' actions              & -27.0837 &   17.651 &   -1.534 & 0.125 &  -61.679 &    7.512    \\
arity                       & -22.9006 &   17.806 &  -1.286 & 0.198 &  -57.800 &   11.999  \\
coupling                    & 0              & nan     & nan       & nan     & nan      & nan      \\
slot-fillable variables     & 0              & nan     & nan       & nan     & nan      & nan      \\
goals                       & -2.53e-16        & 13.455   & -1.88e-17    & 1.000   & -26.372  & 26.372   \\
\hline Granite-3.3-8B-instruct     &                &         &          &         &          &          \\
Pseudo R-squ                & Log-Likelihood &         &          &         &          &          \\
0.0352                      & -99.331        &         &          &         &          &          \\
                            & coef           & std err & z        & $P>|z|$ & {[}0.025 & 0.975{]} \\
`custom' actions                     & 0              & nan     & nan      & nan     & nan      & nan      \\
arity                       & 0              & nan     & nan      & nan     & nan      & nan      \\
coupling                    & -4.953         & 4.509   & -1.099   & 0.272   & -13.79   & 3.884    \\
slot-fillable variables     & -10.9061       & 7.956   & -1.371   & 0.17    & -26.5    & 4.688    \\
goals                       & -16.5726       & 7.71    & -2.15    & 0.032   & -31.684  & -1.461   \\
\hline Llama-3.1-405B-instruct-fp8 &                &         &          &         &          &          \\
Pseudo R-squ                & Log-Likelihood &         &          &         &          &          \\
-0.02809                    & -61.746        &         &          &         &          &          \\
                            & coef           & std err & z        & $P>|z|$ & {[}0.025 & 0.975{]} \\
`custom' actions                     & -4.3217        & 15.636  & -0.276   & 0.782   & -34.968  & 26.325   \\
arity                       & -1.1087        & 16.917  & -0.066   & 0.948   & -34.266  & 32.049   \\
coupling                    & -1.6357        & 5.5     & -0.297   & 0.766   & -12.416  & 9.144    \\
slot-fillable variables     & -15.4646       & 16.811  & -0.92    & 0.358   & -48.414  & 17.485   \\
goals                       & -17.7946       & 10.376  & -1.715   & 0.086   & -38.131  & 2.542    \\
\hline Llama-3-3-70B-instruct      &                &         &          &         &          &          \\
Pseudo R-squ                & Log-Likelihood &         &          &         &          &          \\
0.005184                    & -194.65        &         &          &         &          &          \\
                            & coef           & std err & z        & $P>|z|$ & {[}0.025 & 0.975{]} \\
`custom' actions                     & 0              & nan     & nan      & nan     & nan      & nan      \\
arity                       & 0              & nan     & nan      & nan     & nan      & nan      \\
coupling                    & -4.4632        & 3.174   & -1.406   & 0.16    & -10.684  & 1.757    \\
slot-fillable variables     & 0              & nan     & nan      & nan     & nan      & nan      \\
goals                       & -5.4489        & 3.137   & -1.737   & 0.082   & -11.597  & 0.699    \\
\hline Llama-3.1-8B-instruct       &                &         &          &         &          &          \\
Pseudo R-squ                & Log-Likelihood &         &          &         &          &          \\
-0.03562                    & -106.62        &         &          &         &          &          \\
                            & coef           & std err & z        & $P>|z|$ & {[}0.025 & 0.975{]} \\
`custom' actions                     & -3.6785        & 11.656  & -0.316   & 0.752   & -26.524  & 19.167   \\
arity                       & -19.0184       & 11.842  & -1.606   & 0.108   & -42.228  & 4.191    \\
coupling                    & 0              & nan     & nan      & nan     & nan      & nan      \\
slot-fillable variables     & 0              & nan     & nan      & nan     & nan      & nan      \\
goals                       & -7.0985        & 8.556   & -0.83    & 0.407   & -23.868  & 9.671    \\
\hline Mixtral-8x22B-instruct-v0.1 &                &         &          &         &          &          \\
Pseudo R-squ                & Log-Likelihood &         &          &         &          &          \\
0.09344                     & -91.359        &         &          &         &          &          \\
                            & coef           & std err & z        & $P>|z|$ & {[}0.025 & 0.975{]} \\
`custom' actions                     & 0              & nan     & nan      & nan     & nan      & nan      \\
arity                       & 0              & nan     & nan      & nan     & nan      & nan      \\
coupling                    & -7.5232        & 4.342   & -1.733   & 0.083   & -16.033  & 0.986    \\
slot-fillable variables     & 0              & nan     & nan      & nan     & nan      & nan      \\
goals                       & -27.1984       & 4.446   & -6.118   & 0       & -35.912  & -18.485  \\
\hline Qwen2.5-72B-instruct        &                &         &          &         &          &          \\
Pseudo R-squ                & Log-Likelihood &         &          &         &          &          \\
-0.1284                     & -64.37         &         &          &         &          &          \\
                            & coef           & std err & z        & $P>|z|$ & {[}0.025 & 0.975{]} \\
`custom' actions                     & -18.1791       & 14.71   & -1.236   & 0.217   & -47.01   & 10.652   \\
arity                       & -14.1435       & 15.761  & -0.897   & 0.37    & -45.034  & 16.747   \\
coupling                    & 0              & nan     & nan      & nan     & nan      & nan      \\
slot-fillable variables     & -6.4428        & 16.611  & -0.388   & 0.698   & -39      & 26.114   \\
goals                       & 0              & nan     & nan      & nan     & nan      & nan \\ \hline 
\end{longtable}
\end{center}

\begin{center}
\begin{longtable}{p{3cm}p{1.5cm}p{1.5cm}p{1cm}p{1cm}p{1.5cm}p{1.5cm}}
\caption{Logistic regression analysis for validity of plans for problems with feasible plans with verbose prompt style}
    \label{tab:lrplanvaliditylong} 
    \small
    \\
\hline Deepseek-v3                 &                &         &         &         &          &          \\
Pseudo R-squ                & Log-Likelihood &         &         &         &          &          \\
0.08473                     & -1043.2        &         &         &         &          &          \\
                            & coef           & std err & z       & $P>|z|$ & {[}0.025 & 0.975{]} \\
optimal plan length         & -41.7807       & 5.085   & -8.216  & 0       & -51.748  & -31.814  \\
`custom' actions                     & 62.4231        & 6.135   & 10.174  & 0       & 50.398   & 74.448   \\
arity                       & 0              & nan     & nan     & nan     & nan      & nan      \\
coupling                    & 0              & nan     & nan     & nan     & nan      & nan      \\
slot-fillable variables     & 17.53          & 6.108   & 2.87    & 0.004   & 5.558    & 29.502   \\
goals                       & 0              & nan     & nan     & nan     & nan      & nan      \\
\hline Granite-3.3-8B-instruct     &                &         &         &         &          &          \\
Pseudo R-squ                & Log-Likelihood &         &         &         &          &          \\
-0.0004765                  & -1375.9        &         &         &         &          &          \\
                            & coef           & std err & z       & $P>|z|$ & {[}0.025 & 0.975{]} \\
optimal plan length         & 0              & nan     & nan     & nan     & nan      & nan      \\
`custom' actions                     & 0              & nan     & nan     & nan     & nan      & nan      \\
arity                       & 0              & nan     & nan     & nan     & nan      & nan      \\
coupling                    & 0              & nan     & nan     & nan     & nan      & nan      \\
slot-fillable variables     & 0              & nan     & nan     & nan     & nan      & nan      \\
goals                       & 0              & nan     & nan     & nan     & nan      & nan      \\
\hline Llama-3.1-405B-instruct-fp8 &                &         &         &         &          &          \\
Pseudo R-squ                & Log-Likelihood &         &         &         &          &          \\
0.1782                      & -767.54        &         &         &         &          &          \\
                            & coef           & std err & z       & $P>|z|$ & {[}0.025 & 0.975{]} \\
optimal plan length         & -64.5546       & 5.771   & -11.186 & 0       & -75.866  & -53.243  \\
`custom' actions                     & 95.0417        & 7.206   & 13.19   & 0       & 80.919   & 109.165  \\
arity                       & 0              & nan     & nan     & nan     & nan      & nan      \\
coupling                    & 0              & nan     & nan     & nan     & nan      & nan      \\
slot-fillable variables     & 30.7633        & 6.989   & 4.402   & 0       & 17.066   & 44.461   \\
goals                       & 0              & nan     & nan     & nan     & nan      & nan      \\
\hline Llama-3.3-70B-instruct      &                &         &         &         &          &          \\
Pseudo R-squ                & Log-Likelihood &         &         &         &          &          \\
0.09968                     & -931.42        &         &         &         &          &          \\
                            & coef           & std err & z       & $P>|z|$ & {[}0.025 & 0.975{]} \\
optimal plan length         & -49.7385       & 5.343   & -9.309  & 0       & -60.211  & -39.266  \\
`custom' actions                     & 63.1362        & 6.384   & 9.89    & 0       & 50.624   & 75.649   \\
arity                       & 0              & nan     & nan     & nan     & nan      & nan      \\
coupling                    & 0              & nan     & nan     & nan     & nan      & nan      \\
slot-fillable variables     & 35.2596        & 6.511   & 5.416   & 0       & 22.499   & 48.02    \\
goals                       & 0              & nan     & nan     & nan     & nan      & nan      \\
\hline Llama-3.1-8B-instruct       &                &         &         &         &          &          \\
Pseudo R-squ                & Log-Likelihood &         &         &         &          &          \\
0.003312                    & -1371.4        &         &         &         &          &          \\
                            & coef           & std err & z       & $P>|z|$ & {[}0.025 & 0.975{]} \\
optimal plan length         & -3.4977        & 2.824   & -1.238  & 0.216   & -9.034   & 2.038    \\
`custom' actions                     & 0              & nan     & nan     & nan     & nan      & nan      \\
arity                       & 0              & nan     & nan     & nan     & nan      & nan      \\
coupling                    & 0.375          & 2.823   & 0.133   & 0.894   & -5.157   & 5.907    \\
slot-fillable variables     & 0              & nan     & nan     & nan     & nan      & nan      \\
goals                       & 0              & nan     & nan     & nan     & nan      & nan      \\
\hline Mixtral-8x22B-instruct-v0.1 &                &         &         &         &          &          \\
Pseudo R-squ                & Log-Likelihood &         &         &         &          &          \\
0.02272                     & -1062          &         &         &         &          &          \\
                            & coef           & std err & z       & $P>|z|$ & {[}0.025 & 0.975{]} \\
optimal plan length         & -23.8761       & 5.058   & -4.72   & 0       & -33.79   & -13.962  \\
`custom' actions                     & 50.4787        & 6.101   & 8.274   & 0       & 38.521   & 62.437   \\
arity                       & 0              & nan     & nan     & nan     & nan      & nan      \\
coupling                    & 0              & nan     & nan     & nan     & nan      & nan      \\
slot-fillable variables     & 16.9506        & 6.157   & 2.753   & 0.006   & 4.883    & 29.018   \\
goals                       & 0              & nan     & nan     & nan     & nan      & nan      \\
\hline Qwen2.5-72B-instruct        &                &         &         &         &          &          \\
Pseudo R-squ                & Log-Likelihood &         &         &         &          &          \\
0.03135                     & -1023.5        &         &         &         &          &          \\
                            & coef           & std err & z       & $P>|z|$ & {[}0.025 & 0.975{]} \\
optimal plan length         & -22.2533       & 5.146   & -4.324  & 0       & -32.34   & -12.167  \\
`custom' actions                     & 56.6528        & 6.277   & 9.025   & 0       & 44.349   & 68.956   \\
arity                       & 0              & nan     & nan     & nan     & nan      & nan      \\
coupling                    & 0              & nan     & nan     & nan     & nan      & nan      \\
slot-fillable variables     & 13.7726        & 6.279   & 2.193   & 0.028   & 1.465    & 26.08    \\
goals                       & 0              & nan     & nan     & nan     & nan      & nan     \\ \hline 
\end{longtable}
\end{center}

\begin{center}
\begin{longtable}{p{3cm}p{1.5cm}p{1.5cm}p{1cm}p{1cm}p{1.5cm}p{1.5cm}}
\caption{Logistic regression analysis for validity of plans for problems with feasible plans with concise prompt style}
    \label{tab:lrplanvalidityshort} 
    \small
    \\
\hline Deepseek-v3                 &                &         &         &         &          &          \\
Pseudo R-squ                & Log-Likelihood &         &         &         &          &          \\
0.1384                     & -723.47        &         &         &         &          &          \\
                            & coef           & std err & z       & $P>|z|$ & {[}0.025 & 0.975{]} \\
optimal plan length         & -49.3192 &    5.858 &   -8.419& 0.000&  -60.801&  -37.837  \\
`custom' actions              & 94.7476 &    7.746 &   12.232 & 0.000 &   79.566 &  109.929   \\
arity                       & 0              & nan     & nan     & nan     & nan      & nan      \\
coupling                    & 3.997e-16 &    3.945 & 1.01e-16 & 1.000 &   -7.732 &    7.732      \\
slot-fillable variables     & 26.3420 &    7.395 &    3.562 & 0.000 &   11.848 &   40.836   \\
goals                       & 0              & nan     & nan     & nan     & nan      & nan      \\
\hline Granite-3.3-8B-instruct     &                &         &           &         &          &          \\
Pseudo R-squ                & Log-Likelihood &         &           &         &          &          \\
0.01306                     & -1352.2        &         &           &         &          &          \\
                            & coef           & std err & z         & $P>|z|$ & {[}0.025 & 0.975{]} \\
optimal plan length         & -5.658         & 2.837   & -1.994    & 0.046   & -11.219  & -0.097   \\
`custom' actions                     & 0              & nan     & nan       & nan     & nan      & nan      \\
arity                       & 0              & nan     & nan       & nan     & nan      & nan      \\
coupling                    & -5.5151        & 2.842   & -1.941    & 0.052   & -11.084  & 0.054    \\
slot-fillable variables     & 0              & nan     & nan       & nan     & nan      & nan      \\
goals                       & 0              & nan     & nan       & nan     & nan      & nan      \\
\hline Llama-3.1-405B-instruct-fp8 &                &         &           &         &          &          \\
Pseudo R-squ                & Log-Likelihood &         &           &         &          &          \\
0.1559                      & -713.1         &         &           &         &          &          \\
                            & coef           & std err & z         & $P>|z|$ & {[}0.025 & 0.975{]} \\
optimal plan length         & -56.0808       & 5.887   & -9.527    & 0       & -67.618  & -44.543  \\
`custom' actions                     & 99.9576        & 7.79    & 12.832    & 0       & 84.69    & 115.225  \\
arity                       & 0              & nan     & nan       & nan     & nan      & nan      \\
coupling                    & 1.2189         & 3.931   & 0.31      & 0.757   & -6.486   & 8.924    \\
slot-fillable variables     & 25.7175        & 7.369   & 3.49      & 0       & 11.275   & 40.16    \\
goals                       & 0              & nan     & nan       & nan     & nan      & nan      \\
\hline Llama-3.3-70B-instruct      &                &         &           &         &          &          \\
Pseudo R-squ                & Log-Likelihood &         &           &         &          &          \\
-0.01675                    & -1300          &         &           &         &          &          \\
                            & coef           & std err & z         & $P>|z|$ & {[}0.025 & 0.975{]} \\
optimal plan length         & 0              & nan     & nan       & nan     & nan      & nan      \\
`custom' actions                     & 0              & nan     & nan       & nan     & nan      & nan      \\
arity                       & 0              & nan     & nan       & nan     & nan      & nan      \\
coupling                    & 9.1485         & 3.029   & 3.02      & 0.003   & 3.211    & 15.086   \\
slot-fillable variables     & 8.7886         & 3.01    & 2.92      & 0.003   & 2.89     & 14.687   \\
goals                       & 0              & nan     & nan       & nan     & nan      & nan      \\
\hline Llama-3.1-8B-instruct       &                &         &           &         &          &          \\
Pseudo R-squ                & Log-Likelihood &         &           &         &          &          \\
0.08154                     & -339.31        &         &           &         &          &          \\
                            & coef           & std err & z         & $P>|z|$ & {[}0.025 & 0.975{]} \\
optimal plan length         & -4.00E-16      & 17.191  & -2.32E-17 & 1       & -33.694  & 33.694   \\
`custom' actions                     & -32.2708       & 9.824   & -3.285    & 0.001   & -51.525  & -13.017  \\
arity                       & 0              & nan     & nan       & nan     & nan      & nan      \\
coupling                    & 0              & nan     & nan       & nan     & nan      & nan      \\
slot-fillable variables     & 0              & nan     & nan       & nan     & nan      & nan      \\
goals                       & -99.841        & 20.12   & -4.962    & 0       & -139.276 & -60.406  \\
\hline Mixtral-8x22B-instruct-v0.1 &                &         &           &         &          &          \\
Pseudo R-squ                & Log-Likelihood &         &           &         &          &          \\
-0.008253                   & -1185.3        &         &           &         &          &          \\
                            & coef           & std err & z         & $P>|z|$ & {[}0.025 & 0.975{]} \\
optimal plan length         & 0              & nan     & nan       & nan     & nan      & nan      \\
`custom' actions                     & 0              & nan     & nan       & nan     & nan      & nan      \\
arity                       & 0              & nan     & nan       & nan     & nan      & nan      \\
coupling                    & 0              & nan     & nan       & nan     & nan      & nan      \\
slot-fillable variables     & 29.6337        & 4.892   & 6.057     & 0       & 20.045   & 39.223   \\
goals                       & 8.0143         & 4.786   & 1.674     & 0.094   & -1.366   & 17.395   \\
\hline Qwen2.5-72B-instruct        &                &         &           &         &          &          \\
Pseudo R-squ                & Log-Likelihood &         &           &         &          &          \\
-0.03739                    & -834.25        &         &           &         &          &          \\
                            & coef           & std err & z         & $P>|z|$ & {[}0.025 & 0.975{]} \\
optimal plan length         & -3.29E-15      & 5.972   & -5.51E-16 & 1       & -11.705  & 11.705   \\
`custom' actions                     & 50.0114        & 7.328   & 6.825     & 0       & 35.648   & 64.374   \\
arity                       & 0              & nan     & nan       & nan     & nan      & nan      \\
coupling                    & 4.4832         & 4.009   & 1.118     & 0.263   & -3.374   & 12.341   \\
slot-fillable variables     & 19.8769        & 7.324   & 2.714     & 0.007   & 5.522    & 34.232   \\
goals                       & 0              & nan     & nan       & nan     & nan      & nan     \\ \hline 
\end{longtable}
\end{center}

\begin{center}
\begin{longtable}{p{3cm}p{1.5cm}p{1.5cm}p{1cm}p{1cm}p{1.5cm}p{1.5cm}}
\caption{Logistic regression analysis for optimality of plans for problems with feasible plans with verbose prompt style}
    \label{tab:lrplanoptimallong} 
    \small
    \\
\hline Deepseek-v3                 &                &         &        &         &          &          \\
Pseudo R-squ                & Log-Likelihood &         &        &         &          &          \\
0.02228                     & -1342.1        &         &        &         &          &          \\
                            & coef           & std err & z      & $P>|z|$ & {[}0.025 & 0.975{]} \\
optimal plan length         & -1.8931        & 4.346   & -0.436 & 0.663   & -10.411  & 6.625    \\
`custom' actions                     & 0              & nan     & nan    & nan     & nan      & nan      \\
arity                       & 0              & nan     & nan    & nan     & nan      & nan      \\
coupling                    & -16.0413       & 3.061   & -5.24  & 0       & -22.041  & -10.041  \\
slot-fillable variables     & 7.9159         & 4.59    & 1.725  & 0.085   & -1.08    & 16.912   \\
goals                       & 0              & nan     & nan    & nan     & nan      & nan      \\
\hline Granite-3.3-8B-instruct     &                &         &        &         &          &          \\
Pseudo R-squ                & Log-Likelihood &         &        &         &          &          \\
0.04275                     & -1236.8        &         &        &         &          &          \\
                            & coef           & std err & z      & $P>|z|$ & {[}0.025 & 0.975{]} \\
optimal plan length         & 0              & nan     & nan    & nan     & nan      & nan      \\
`custom' actions                     & -2.7211        & 3.102   & -0.877 & 0.38    & -8.801   & 3.359    \\
arity                       & 0              & nan     & nan    & nan     & nan      & nan      \\
coupling                    & -28.8463       & 3.278   & -8.799 & 0       & -35.272  & -22.421  \\
slot-fillable variables     & 0              & nan     & nan    & nan     & nan      & nan      \\
goals                       & 0              & nan     & nan    & nan     & nan      & nan      \\
\hline Llama-3.1-405B-instruct-fp8 &                &         &        &         &          &          \\
Pseudo R-squ                & Log-Likelihood &         &        &         &          &          \\
0.05037                     & -1236.7        &         &        &         &          &          \\
                            & coef           & std err & z      & $P>|z|$ & {[}0.025 & 0.975{]} \\
optimal plan length         & -6.761         & 4.742   & -1.426 & 0.154   & -16.055  & 2.533    \\
`custom' actions                     & 22.0291        & 5.675   & 3.882  & 0       & 10.906   & 33.152   \\
arity                       & 0              & nan     & nan    & nan     & nan      & nan      \\
coupling                    & -22.9507       & 3.236   & -7.093 & 0       & -29.292  & -16.609  \\
slot-fillable variables     & 21.0509        & 5.726   & 3.676  & 0       & 9.828    & 32.274   \\
goals                       & 0              & nan     & nan    & nan     & nan      & nan      \\
\hline Llama-3.3-70B-instruct      &                &         &        &         &          &          \\
Pseudo R-squ                & Log-Likelihood &         &        &         &          &          \\
0.07216                     & -1209.5        &         &        &         &          &          \\
                            & coef           & std err & z      & $P>|z|$ & {[}0.025 & 0.975{]} \\
optimal plan length         & -25.7043       & 6.722   & -3.824 & 0       & -38.88   & -12.529  \\
`custom' actions                     & 11.1087        & 5.649   & 1.966  & 0.049   & 0.037    & 22.181   \\
arity                       & 18.4486        & 6.866   & 2.687  & 0.007   & 4.992    & 31.905   \\
coupling                    & -18.039        & 3.248   & -5.554 & 0       & -24.405  & -11.673  \\
slot-fillable variables     & 28.5103        & 5.866   & 4.86   & 0       & 17.013   & 40.007   \\
goals                       & 0              & nan     & nan    & nan     & nan      & nan      \\
\hline Llama-3.1-8B-instruct       &                &         &        &         &          &          \\
Pseudo R-squ                & Log-Likelihood &         &        &         &          &          \\
0.0603                      & -1023.4        &         &        &         &          &          \\
                            & coef           & std err & z      & $P>|z|$ & {[}0.025 & 0.975{]} \\
optimal plan length         & 0              & nan     & nan    & nan     & nan      & nan      \\
`custom' actions                     & -28.2565       & 3.368   & -8.39  & 0       & -34.858  & -21.655  \\
arity                       & 0              & nan     & nan    & nan     & nan      & nan      \\
coupling                    & -30.1313       & 3.653   & -8.248 & 0       & -37.291  & -22.971  \\
slot-fillable variables     & 0              & nan     & nan    & nan     & nan      & nan      \\
goals                       & 0              & nan     & nan    & nan     & nan      & nan      \\
\hline Mixtral-8x22B-instruct-v0.1 &                &         &        &         &          &          \\
Pseudo R-squ                & Log-Likelihood &         &        &         &          &          \\
0.04701                     & -1302.3        &         &        &         &          &          \\
                            & coef           & std err & z      & $P>|z|$ & {[}0.025 & 0.975{]} \\
optimal plan length         & -0.3057        & 6.459   & -0.047 & 0.962   & -12.965  & 12.354   \\
`custom' actions                     & 0              & nan     & nan    & nan     & nan      & nan      \\
arity                       & 17.5519        & 6.677   & 2.629  & 0.009   & 4.465    & 30.638   \\
coupling                    & -21.6725       & 3.125   & -6.935 & 0       & -27.798  & -15.547  \\
slot-fillable variables     & 5.1429         & 4.821   & 1.067  & 0.286   & -4.306   & 14.592   \\
goals                       & 0              & nan     & nan    & nan     & nan      & nan      \\
\hline Qwen2.5-72B-instruct        &                &         &        &         &          &          \\
Pseudo R-squ                & Log-Likelihood &         &        &         &          &          \\
0.05514                     & -1299.6        &         &        &         &          &          \\
                            & coef           & std err & z      & $P>|z|$ & {[}0.025 & 0.975{]} \\
optimal plan length         & 0              & nan     & nan    & nan     & nan      & nan      \\
`custom' actions                     & 0              & nan     & nan    & nan     & nan      & nan      \\
arity                       & 0              & nan     & nan    & nan     & nan      & nan      \\
coupling                    & -24.8377       & 3.085   & -8.051 & 0       & -30.884  & -18.791  \\
slot-fillable variables     & 9.6276         & 3.015   & 3.194  & 0.001   & 3.719    & 15.536   \\
goals                       & 0              & nan     & nan    & nan     & nan      & nan   \\ \hline   
\end{longtable}
\end{center}

\begin{center}
\begin{longtable}{p{3cm}p{1.5cm}p{1.5cm}p{1cm}p{1cm}p{1.5cm}p{1.5cm}}
\caption{Logistic regression analysis for optimality of plans for problems with feasible plans with concise prompt style}
    \label{tab:lrplanoptimalshort} 
    \small
    \\*
\hline Deepseek-v3                 &                &         &        &         &          &          \\
Pseudo R-squ                & Log-Likelihood &         &        &         &          &          \\
0.009851                     & -1279.7        &         &        &         &          &          \\
                            & coef           & std err & z      & $P>|z|$ & {[}0.025 & 0.975{]} \\
optimal plan length         & 0              & nan     & nan    & nan     & nan      & nan    \\
`custom' actions              & 20.4360 &    4.515 &    4.527 & 0.000 &   11.588 &   29.284      \\
arity                       & 4.1622 &    4.290 &    0.970 & 0.332 &   -4.246 &   12.570      \\
coupling                    & -22.6719 &    3.180 &   -7.130 & 0.000 &  -28.904 &  -16.440  \\
slot-fillable variables     & 0              & nan     & nan    & nan     & nan      & nan   \\
goals                       & 0              & nan     & nan    & nan     & nan      & nan      \\
\hline Granite-3.3-8B-instruct     &                &         &           &         &          &          \\
Pseudo R-squ                & Log-Likelihood &         &           &         &          &          \\
0.01306                     & -1352.2        &         &           &         &          &          \\
                            & coef           & std err & z         & $P>|z|$ & {[}0.025 & 0.975{]} \\
optimal plan length         & -5.658         & 2.837   & -1.994    & 0.046   & -11.219  & -0.097   \\
`custom' actions                     & 0              & nan     & nan       & nan     & nan      & nan      \\
arity                       & 0              & nan     & nan       & nan     & nan      & nan      \\
coupling                    & -5.5151        & 2.842   & -1.941    & 0.052   & -11.084  & 0.054    \\
slot-fillable variables     & 0              & nan     & nan       & nan     & nan      & nan      \\
goals                       & 0              & nan     & nan       & nan     & nan      & nan      \\
\hline Llama-3.1-405B-instruct-fp8 &                &         &           &         &          &          \\
Pseudo R-squ                & Log-Likelihood &         &           &         &          &          \\
0.1559                      & -713.1         &         &           &         &          &          \\
                            & coef           & std err & z         & $P>|z|$ & {[}0.025 & 0.975{]} \\
optimal plan length         & -56.0808       & 5.887   & -9.527    & 0       & -67.618  & -44.543  \\
`custom' actions                     & 99.9576        & 7.79    & 12.832    & 0       & 84.69    & 115.225  \\
arity                       & 0              & nan     & nan       & nan     & nan      & nan      \\
coupling                    & 1.2189         & 3.931   & 0.31      & 0.757   & -6.486   & 8.924    \\
slot-fillable variables     & 25.7175        & 7.369   & 3.49      & 0       & 11.275   & 40.16    \\
goals                       & 0              & nan     & nan       & nan     & nan      & nan      \\
\hline Llama-3.3-70B-instruct      &                &         &           &         &          &          \\
Pseudo R-squ                & Log-Likelihood &         &           &         &          &          \\
-0.01675                    & -1300          &         &           &         &          &          \\
                            & coef           & std err & z         & $P>|z|$ & {[}0.025 & 0.975{]} \\
optimal plan length         & 0              & nan     & nan       & nan     & nan      & nan      \\
`custom' actions                     & 0              & nan     & nan       & nan     & nan      & nan      \\
arity                       & 0              & nan     & nan       & nan     & nan      & nan      \\
coupling                    & 9.1485         & 3.029   & 3.02      & 0.003   & 3.211    & 15.086   \\
slot-fillable variables     & 8.7886         & 3.01    & 2.92      & 0.003   & 2.89     & 14.687   \\
goals                       & 0              & nan     & nan       & nan     & nan      & nan      \\
\hline Llama-3.1-8B-instruct       &                &         &           &         &          &          \\
Pseudo R-squ                & Log-Likelihood &         &           &         &          &          \\
0.08154                     & -339.31        &         &           &         &          &          \\
                            & coef           & std err & z         & $P>|z|$ & {[}0.025 & 0.975{]} \\
optimal plan length         & -4.00E-16      & 17.191  & -2.32E-17 & 1       & -33.694  & 33.694   \\
`custom' actions                     & -32.2708       & 9.824   & -3.285    & 0.001   & -51.525  & -13.017  \\
arity                       & 0              & nan     & nan       & nan     & nan      & nan      \\
coupling                    & 0              & nan     & nan       & nan     & nan      & nan      \\
slot-fillable variables     & 0              & nan     & nan       & nan     & nan      & nan      \\
goals                       & -99.841        & 20.12   & -4.962    & 0       & -139.276 & -60.406  \\
\hline Mixtral-8x22B-instruct-v0.1 &                &         &           &         &          &          \\
Pseudo R-squ                & Log-Likelihood &         &           &         &          &          \\
-0.008253                   & -1185.3        &         &           &         &          &          \\
                            & coef           & std err & z         & $P>|z|$ & {[}0.025 & 0.975{]} \\
optimal plan length         & 0              & nan     & nan       & nan     & nan      & nan      \\
`custom' actions                     & 0              & nan     & nan       & nan     & nan      & nan      \\
arity                       & 0              & nan     & nan       & nan     & nan      & nan      \\
coupling                    & 0              & nan     & nan       & nan     & nan      & nan      \\
slot-fillable variables     & 29.6337        & 4.892   & 6.057     & 0       & 20.045   & 39.223   \\
goals                       & 8.0143         & 4.786   & 1.674     & 0.094   & -1.366   & 17.395   \\
\hline Qwen2.5-72B-instruct        &                &         &           &         &          &          \\
Pseudo R-squ                & Log-Likelihood &         &           &         &          &          \\
-0.03739                    & -834.25        &         &           &         &          &          \\
                            & coef           & std err & z         & $P>|z|$ & {[}0.025 & 0.975{]} \\
optimal plan length         & -3.29E-15      & 5.972   & -5.51E-16 & 1       & -11.705  & 11.705   \\
`custom' actions                     & 50.0114        & 7.328   & 6.825     & 0       & 35.648   & 64.374   \\
arity                       & 0              & nan     & nan       & nan     & nan      & nan      \\
coupling                    & 4.4832         & 4.009   & 1.118     & 0.263   & -3.374   & 12.341   \\
slot-fillable variables     & 19.8769        & 7.324   & 2.714     & 0.007   & 5.522    & 34.232   \\
goals                       & 0              & nan     & nan       & nan     & nan      & nan     \\ \hline 
\end{longtable}
\end{center}

\subsection*{LLM Plan Length Distributions}

\begin{figure}[htbp]
\centering
  \includegraphics[scale=0.34]{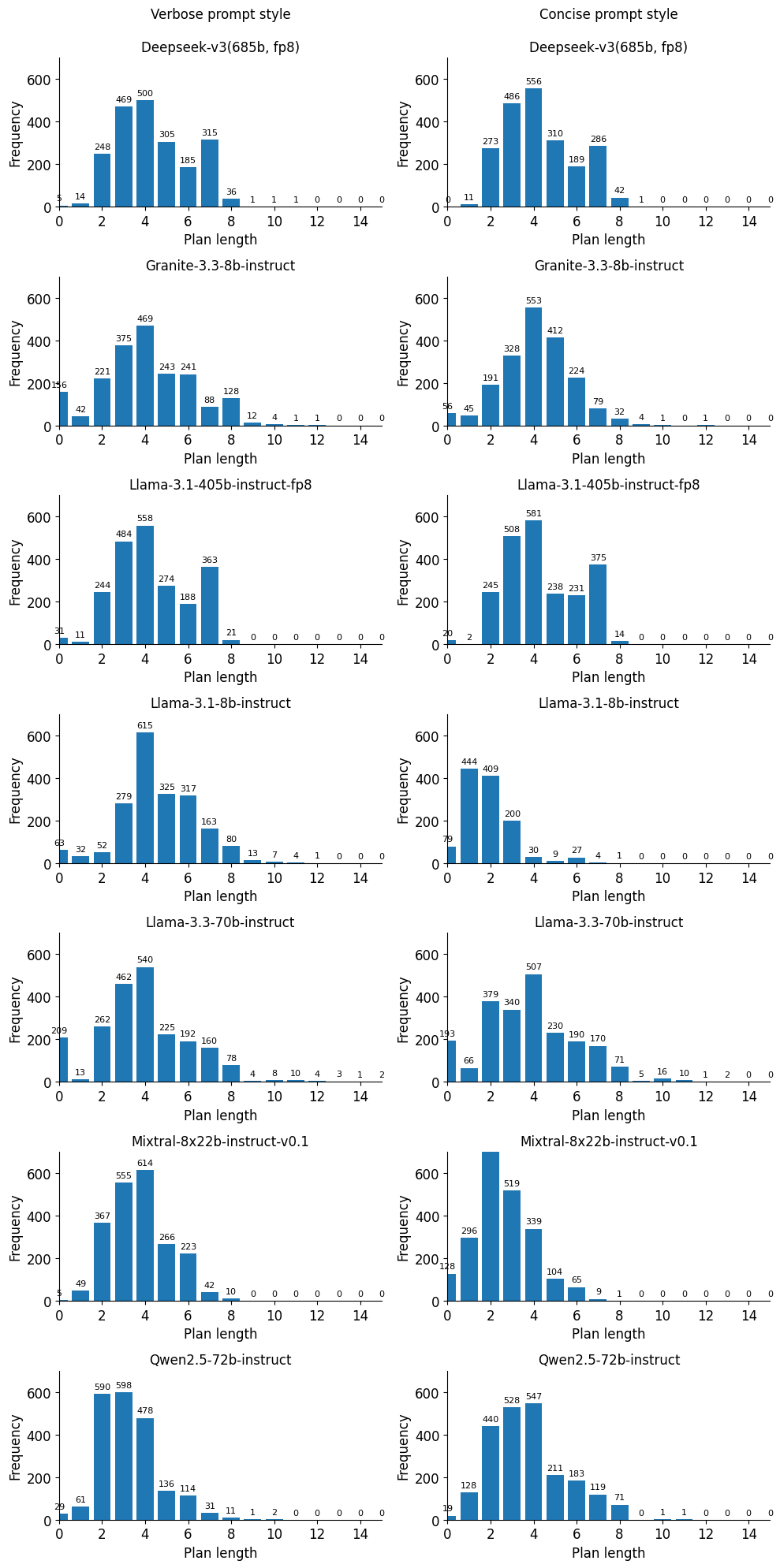}
  \caption{Distribution of LLM-generated plan lengths for different prompt styles}
  \label{fig:planlengthverbose}
\end{figure}

\begin{figure}[htbp]
\centering
  \includegraphics[scale=0.6]{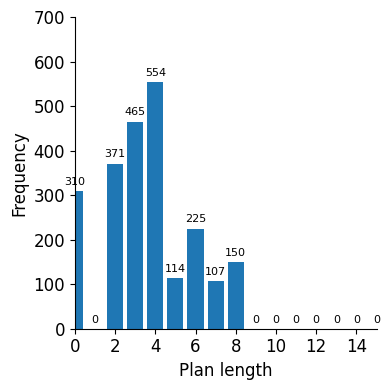}
  \caption{Distribution of optimal plan lengths computed by the k$\ast$ planner}
  \label{fig:planlengthconcise}
\end{figure}

\end{document}